%% file: main.tex
\documentclass[11pt]{article}

\usepackage[preprint]{acl}

\usepackage{times}
\usepackage{latexsym}

\usepackage[T1]{fontenc}

\usepackage{microtype}

\usepackage{inconsolata}

\usepackage{graphicx}

\usepackage{amsmath}
\usepackage{amssymb}
\usepackage{booktabs}
\usepackage{tabularx}
\usepackage{makecell}
\usepackage{enumitem}
\usepackage{threeparttable}
\usepackage{siunitx}
\usepackage[UTF8, scheme=plain]{ctex}

\title{Structured Style-Rewrite with Chain-of-Thought Planning for Low-Resource Character Dialogue}

\author{
	Chanhui Zhu \\
	Guangdong University of Finance \\
	\texttt{241543255@m.gduf.edu.cn}
}

\begin{document}
\maketitle

\begin{abstract}
Applying Small Language Models (SLMs) to Chinese character-driven generation remains challenging
due to data scarcity and the difficulty of disentangling character style.
Standard Supervised Fine-Tuning (SFT) often captures surface-level semantics
but produces frequent ``Out-Of-Character'' (OOC) outputs.
We frame this as a controlled sentence-level style rewriting task,
which isolates stylistic quality from dialogue context management.
We propose a Structured Style-Rewrite Framework that decomposes character style
into interpretable format signature, syntactic, and pragmatic dimensions,
combined with Chain-of-Thought (CoT) supervision for explicit style planning.
A CoT-Shared Direct Preference Optimization (DPO) stage further aligns style planning with surface realization
by ensuring preference learning targets output-level style execution rather than reasoning trace differences.
Experiments across eight characters from four diverse source domains demonstrate that our method
enables a Qwen3-1.7B model to achieve a Valid Style Score of $0.632$
while maintaining strong semantic fidelity (0.878),
placing on the Pareto frontier among the evaluated systems
and outperforming significantly larger baselines (e.g., GLM-4.7) on consumer hardware.
\end{abstract}

\section{Introduction}
\input{sections/introduction}

\section{Related Work}
\input{sections/related_work}

\section{Method}
\input{sections/method}

\section{Experiments}
\input{sections/experiments}

\section{Conclusion}
\input{sections/conclusion}

\section*{Limitations}
\input{sections/limitations}

\section*{Ethical Considerations}
This work focuses exclusively on fictional character style modeling.
The framework is not designed for impersonating real individuals; all training characters are publicly available fictional personas.
The synthetic data pipeline (Appendix~\ref{app:data_processing}) includes explicit content filtering to remove harmful, violent, or personally identifiable content.
Data sources comply with their respective licenses and ACL data usage policies (Appendix~\ref{app:data_ethics}).
We acknowledge the theoretical risk of misuse for persona impersonation, but note that the low-resource setting and character-specific style vectors substantially limit transferability to arbitrary real-world targets.

\bibliography{anthology,custom}

\appendix
\input{sections/appendixes}

\end{document}

%% file: sections/introduction.tex
Controllable dialogue generation is a central problem in generative AI.
Character-driven applications---such as animation, games, and fictional narratives---require both consistent stylistic identity and faithful content preservation,
a requirement we study in the controlled setting of sentence-level style rewriting.
Despite recent advances, existing dialogue models still struggle to reproduce
\emph{consistent and distinctive character styles},
especially under low-resource conditions.

This difficulty stems from two fundamental challenges.
First, character style is inherently high-dimensional,
involving lexical preferences, syntactic patterns, pragmatic tendencies,
and idiosyncratic speech habits.
Second, most fictional characters exist in extremely low-resource settings,
where only a small number of utterances are available.
While prior work on text style transfer has shown promising results
in controlling attributes such as sentiment or register
\cite{gao2019latent,hu2021controllable,smith2020style},
most approaches rely on holistic latent embeddings
or large annotated datasets,
limiting their interpretability and effectiveness
for low-resource, fine-grained character modeling.
Moreover, prompt-based role-playing with large language models
often suffers from stylistic instability
and frequent out-of-character (OOC) behaviors.

\begin{figure}[t]
	\centering
	\includegraphics[width=\linewidth]{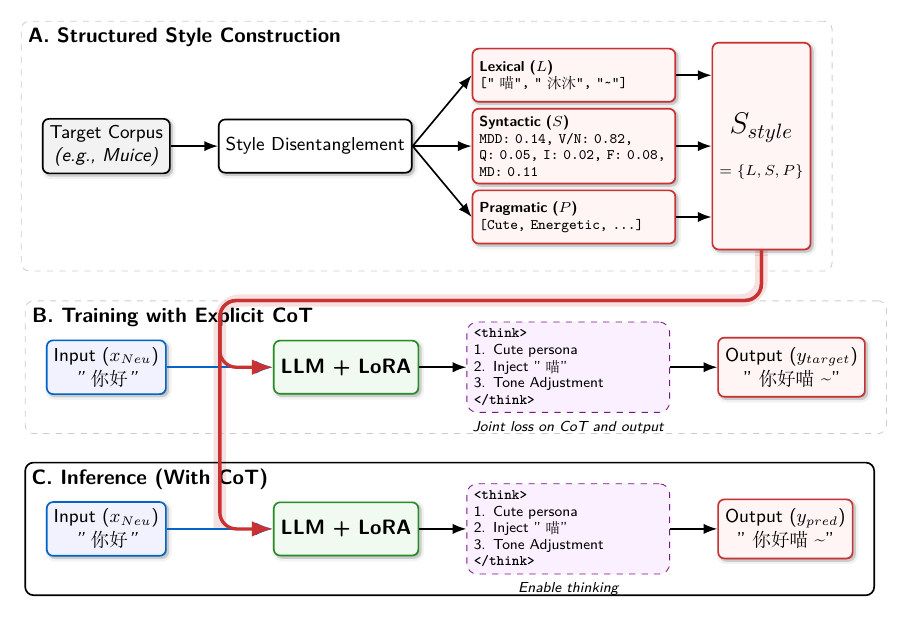}
	\caption{Overview of the proposed structured style modeling and
	style-conditioned generation framework.}
	\label{fig:framework_overview}
\end{figure}

To address these challenges, we propose a Structured Style-Rewrite Framework
that decomposes character style into
format signature (TF-PMI keywords and punctuation), syntactic (six statistical dimensions),
and pragmatic components,
and employs a two-stage training pipeline
(SFT + CoT-Shared DPO) to bridge style planning and surface realization (Figure~\ref{fig:framework_overview}).

\paragraph{Contributions.}
Our main contributions are summarized as follows:

\begin{itemize}
  \item \textbf{Structured Style Representation.} 
    We decompose character style into interpretable format signature, syntactic, 
    and pragmatic components unified into a compositional style vector, 
    enabling transparent conditioning under low-resource constraints.
  \item \textbf{CoT-Shared DPO.} 
    We introduce CoT supervision for explicit style planning and a 
    CoT-Shared DPO stage that concentrates preference gradients on 
    output-level execution, preventing reasoning-trace shortcuts.
  \item \textbf{Empirical Validation.} 
    Experiments across eight characters demonstrate Pareto-optimal 
    style–semantic balance (VSS 0.632, Semantic 0.878), with few-shot 
    adaptation (N=25) to unseen characters under cold-start conditions.
\end{itemize}

%% file: sections/related_work.tex
\subsection{Style-Controlled and Persona-Based Dialogue Generation}

Controlling stylistic attributes in dialogue generation
has been extensively studied in both text style transfer
and persona-based dialogue systems.
Early work focused on learning latent style representations
or disentangling content and style
\cite{gao2019latent,hu2021controllable}.
In persona-based dialogue,
style or persona traits are typically encoded
as holistic embeddings or discrete attribute labels
\cite{smith2020style}.
Recent work also explores low-dimensional style representations
for disentangling stylistic factors
\cite{khalid2025slimllms}.

These approaches generally rely on large annotated datasets or opaque representations,
limiting their applicability to low-resource character settings with fine-grained style requirements.

\citet{tang2025thinkingcharacteradvancingroleplaying} propose Role-Aware Reasoning for multi-turn role-playing dialogue;
our work complements theirs by focusing on low-resource sentence-level rewriting with static structured style vectors.

\subsection{Retrieval-Based and Rewrite-Based Dialogue Systems}

Retrieval-based dialogue systems select responses
by matching the dialogue context with candidate utterances
from a corpus
\cite{yan2016learning,wu-etal-2017-sequential,zhou-etal-2018-multi}.
Recent retrieval-augmented generation (RAG) frameworks
further integrate external knowledge
to improve factuality and relevance
\cite{xu-etal-2022-beyond,yu2024rag}.
Several studies have also explored retrieval-based
or rewrite-based strategies
for character-style dialogue generation
\cite{han2022character,fu2021stylistic}.
In non-parallel author-style rewriting,
the DRAG framework introduces explicit attribute control
through a director-generator design
\cite{singh-etal-2021-drag},
which is closely related to our rewrite-based setting.

\subsection{Few-Shot and Low-Resource Style Learning}

With the increasing capabilities of large language models,
few-shot learning has become a common paradigm
for dialogue style transfer
\cite{roy2023fewshot,krishna-etal-2022-shot}.
Techniques such as pseudo dialogue prompting
retrieve relevant character histories
to improve stylistic consistency
under limited supervision
\cite{han2022character}.

\citet{krishna-etal-2022-shot}
further study low-resource multilingual style transfer
with controllable few-shot conditioning.
Compared with example-driven approaches, our method emphasizes explicit structured style vectors for improved interpretability under sparse data.

%% file: sections/method.tex
\subsection{Task Formulation}

Given a neutral sentence $x$ and a target style vector $S$, the character style transfer task aims to generate a stylized sentence:

\begin{equation}
	y = f(x, S)
\end{equation}

The generated sentence $y$ must preserve the semantic proposition of $x$ (content consistency), conform to the stylistic constraints specified by $S$ (style consistency), and remain grammatically fluent.

This task corresponds to a rewrite-based variant of text style transfer \cite{gao2019latent,hu2021controllable,smith2020style}, where the objective is to rewrite the input sentence into a target character style while preserving its original content.

The style input $S$ is defined as a structured vector, formally represented as:

\begin{equation}
	S = [\text{Format Signature}, \text{Syntactic}, \text{Pragmatic}]
\end{equation}

Each component of the structured style vector $S$ can be further decomposed into multiple interpretable sub-dimensions. A brief definition of the style vector is shown in Table~\ref{tab:style_vector}.

\begin{table}[t]
	\centering
	\small
	\setlength{\tabcolsep}{3pt}
	\caption{Structure of the style input vector $S$.}
	\label{tab:style_vector}
	\begin{tabular}{@{}p{0.30\columnwidth}p{0.18\columnwidth}p{0.45\columnwidth}@{}}
		\toprule
		\textbf{Component} & \textbf{Dim.} & \textbf{Description} \\
		\midrule
		Format Signature & $\lvert\mathcal{V}\rvert + \lvert\mathcal{P}\rvert$ &
		TF--PMI-ranked character-specific keywords and punctuation tendencies \\
		Syntactic & $6$ &
		Statistical syntactic style vector \\
		Pragmatic & $50$ &
		Multi-label style distribution predicted by the refiner \\
		\bottomrule
	\end{tabular}
\end{table}

Detailed descriptions and examples of these dimensions are provided in Appendix~\ref{app:style_vector}.

The structured vector $S$ can be instantiated either as a set of numerical vectors or as a collection of control prompts that constrain stylistic generation. These two representations are introduced separately and applied in different experimental settings in the subsequent sections.

\subsection{Style Representation and Extraction}

\paragraph{Style Format Signature Construction.}

Inspired by \citet{roy2023fewshot}, we extract character-specific keywords
via a TF-weighted PMI scheme with cross-corpus balanced normalization (see Appendix~\ref{app:tfpmi}).

The resulting TF--PMI scores construct the lexical subcomponent of the format-signature component of $S$.

For punctuation tendencies, we compute turn-level activity rates of composite punctuation patterns
(e.g., multi-``!''/``?'', interrobang, ellipsis, repeated tildes) together with special symbols and emoji,
and retain active items above a fixed threshold as the punctuation signature list.

\paragraph{Syntactic Style Features.}

Rather than relying on constituency parse-based PCFG statistics,
we adopt six interpretable statistical dimensions
that directly capture the surface-level rhythmic and structural
habits of character speech. Specifically,
we compute: (1) MDD (mean dependency distance, reflecting structural complexity),
(2) verb--noun ratio (action vs.\ description propensity), (3) question ratio (fraction of interrogative utterances),
(4) imperative ratio (fraction of imperative constructions), (5) fragment ratio (colloquial compression),
and (6) marker density (discourse particle frequency).

These dimensions are computed directly from dependency parses
of the character corpus using HanLP~\cite{he-choi-2021-stem},
yielding a compact $v_{\text{syn}} \in \mathbb{R}^{6}$
that captures dominant structural tendencies
while preserving interpretability.
Formal definitions and per-character values
are provided in Appendix~\ref{app:statistical_vector}.

\paragraph{Pragmatic Style Modeling.}

While lexical and syntactic features capture surface-level style markers,
pragmatic tendencies (personality traits, emotional tone, social stance)
determine \emph{how} a character expresses meaning beyond word choice and sentence structure.
Rather than sequentially predicting pragmatic styles---which may propagate errors---a style profile is constructed as a soft conditioning signal encoding the global distribution of pragmatic tendencies inferred from the character corpus.
As the ablation study confirms (\S\ref{sec:ablation}), pragmatic features provide fine-grained refinements ($\Delta = -0.025$) that complement the dominant lexical component rather than acting as an independent primary driver of style transfer.

Initial pragmatic labels are obtained via clustering-based heuristics on character utterances,
which may introduce label noise under low-resource conditions.
To improve reliability, we train a Context-Aware Style Refiner (Appendix~\ref{app:style_refiner}) that corrects pseudo-labels via prototype-guided refinement.

\subsection{Conditional Style Transfer}
\label{sec:conditional_style_transfer}

\subsubsection{Conditional LoRA with Style Prefix Injection}

We adopt Low-Rank Adaptation (LoRA) as the basis for parameter-efficient fine-tuning.
Instead of directly modifying the backbone model parameters,
we inject structured style information through a learnable \emph{style prefix}.

\paragraph{Style Encoder.}
We introduce a lightweight neural encoder, denoted as \textit{StyleEncoder},
which maps a structured syntactic style vector into the hidden representation space
of the backbone language model.
Given a syntactic style vector $s_{\text{syn}} \in \mathbb{R}^{6}$,
the encoder produces a style embedding
$
	\mathbf{e}_{\text{style}} = \text{StyleEncoder}(s_{\text{syn}}),
$

where $\mathbf{e}_{\text{style}}$ matches the model hidden dimension.

\paragraph{Prefix Injection.}
During the forward pass, the style embedding is prepended to the token embeddings
of the input sequence, forming a soft prompt that conditions generation.
This prefix is masked from the language modeling loss,
ensuring that it serves solely as a conditioning signal rather than a generation target.

By injecting style information as a prefix rather than as textual tokens,
the model is encouraged to internalize structured stylistic constraints
within its latent representations.

\subsubsection{Instruction-Based Input Formulation}

We formulate the task using an explicit instructional prompt containing character identity,
stylistic profiles, and the neutral input (see Appendix~\ref{app:prompt_templates} for prompt templates).

This formulation reframes style transfer as an explicit transformation task
rather than a conversational response,
thereby reducing semantic leakage and stabilizing training.

\subsubsection{Chain-of-Thought Supervision}

To enable the model to handle styles that involve semantic inversion or implicit reasoning,
we incorporate chain-of-thought (CoT) supervision during training.
Each training target consists of a reasoning trace followed by the final stylized output.

This design encourages the model to explicitly reason about
how stylistic constraints interact with semantic content,
rather than relying on shallow pattern matching.
The language modeling objective is applied jointly to both the reasoning trace
and the final response.

\subsubsection{Training Objective}

The overall training objective applies hierarchical weighting
to the language modeling losses over the CoT reasoning trace
and the final stylized output:

\begin{equation}
	L_{\text{total}} = \lambda_{\text{CoT}} \cdot L_{\text{CoT}}
	+ \lambda_{\text{out}} \cdot L_{\text{out}},
\end{equation}

where $L_{\text{CoT}}$ is the standard autoregressive loss
over the \texttt{<think>} reasoning segment,
$L_{\text{out}}$ is the autoregressive loss
over the final stylized response,
and we set $\lambda_{\text{CoT}} = 0.7$ and $\lambda_{\text{out}} = 1.0$,
where the CoT weight balances supervision effectiveness against overfitting risk (see Appendix~\ref{app:cot_dynamics} for a training dynamics analysis).

\subsubsection{Preference-Based Style Alignment}
\label{sec:dpo}

Although SFT with CoT supervision establishes
reliable style planning, we observe a persistent gap
between the model's internal reasoning and its surface realization:
the model occasionally identifies correct stylistic strategies
in its CoT trace but fails to execute them in the final output
(see Appendix~\ref{app:error_analysis} for examples).
To bridge this gap, we introduce a second training stage
based on Direct Preference Optimization~\cite{rafailov2024directpreferenceoptimizationlanguage}.

\paragraph{Hard Negative Construction.}
To construct preference pairs, we generate rejected samples
by removing the Lexical Signature from the conditioning prompt
and re-running inference with the SFT model.
This produces outputs that are semantically faithful
but stylistically bland---precisely the hard negative
that forces DPO to learn lexical integration
rather than semantic shortcuts.
Pairs with a length ratio outside $(0.6, 1.5)$ are filtered
to prevent length-based reward hacking.

\paragraph{CoT-Shared DPO.}
Standard DPO applied to CoT-augmented models
suffers from a \emph{CoT shortcut}:
the model discriminates chosen and rejected samples
by exploiting differences in the reasoning trace
rather than learning from output-level style execution.
To prevent this, we share the chosen sample's
\texttt{<think>} content across both chosen and rejected pairs.
Because the CoT segment is identical,
its gradient contributions cancel to zero,
forcing DPO gradients to concentrate on the output segment.

\paragraph{Training Configuration.}
We apply DPO to the SFT checkpoint (Model(SFT))
with $\beta = 0.1$ and a learning rate of $5 \times 10^{-7}$.
The final checkpoint is selected at Step 150
based on validation style metrics.



%% file: sections/experiments.tex
This section evaluates the proposed lightweight role-playing framework in terms of controllability, robustness, and semantic fidelity under style transfer settings. The evaluation covers automatic metrics, LLM-based judgment, and human assessment.

\subsection{Experimental Setup}
\label{sec:experiments}

\paragraph{Datasets.}
We evaluate on Chinese dialogue datasets spanning diverse domains.
The training corpus contains 8,272 synthetic pairs before oversampling
(9,521 after), across eight characters with a strict 1:1 neutral-to-style ratio.
Per-character details are in Appendix~\ref{app:setup} (Table~\ref{tab:data_scale}).

\paragraph{Baselines.}
We compare against three representative paradigms spanning retrieval, multi-task SFT, and strong LLM prompting, as summarized in Table~\ref{tab:baseline_summary}.
Baseline C uses GLM-4.7\footnote{GLM-4.7 is a model within the GLM-4.5 ARC family~\cite{5team2025glm45agenticreasoningcoding}.} as the prompting backbone.

\begin{table}[t]
	\centering
	\small
	\setlength{\tabcolsep}{3pt}
	\caption{Baseline models and proposed variants.}
	\label{tab:baseline_summary}
	\begin{tabular}{@{}l@{\hspace{2pt}}l@{\hspace{2pt}}l@{}}
		\toprule
		Label & Paradigm & Backbone \\
		\midrule
		Baseline A & RAG + Few-shot & Qwen3-1.7B \\
		Baseline B & Vanilla SFT & Qwen3-1.7B \\
		Baseline C & Strong LLM Prompting & GLM-4.7 \\
		\midrule
		Ours(SFT) & Structured SFT + CoT & Qwen3-1.7B \\
		Ours(DPO) & SFT + CoT-Shared DPO & Qwen3-1.7B \\
		\bottomrule
	\end{tabular}
\end{table}

\paragraph{Metrics.}
Automatic evaluation uses four metrics: \emph{Semantic Score} (cosine similarity via \texttt{BGE-large-zh-v1.5}), \emph{Style Score} (cosine similarity to the target character's centroid from a RoBERTa-based style discriminator), \emph{Valid Style Score} ($S_{\text{valid}} = S_{\text{raw}} \times \mathbb{I}(\text{Semantic} > 0.75)$), and \emph{H-Score} (harmonic mean of Semantic and Style, measuring their joint optimization).
The style discriminator uses \texttt{chinese-roberta-wwm-ext}~\cite{cui2021pretraining} with a 256-dimensional projection head, trained on a character-stratified 80/20 split of the stylized training utterances; centroids are computed exclusively from the training split and remain frozen during evaluation. Full training hyperparameters are listed in Appendix~\ref{sec:refiner_metrics}.
Human evaluation uses single-response absolute scoring on a 1--5 Likert scale along three dimensions (see Section~\ref{sec:human_eval}).

\subsection{Main Results}

\subsubsection{Automatic Evaluation Results}

\begin{figure}[t]
	\centering
	\includegraphics[width=1.00\linewidth]{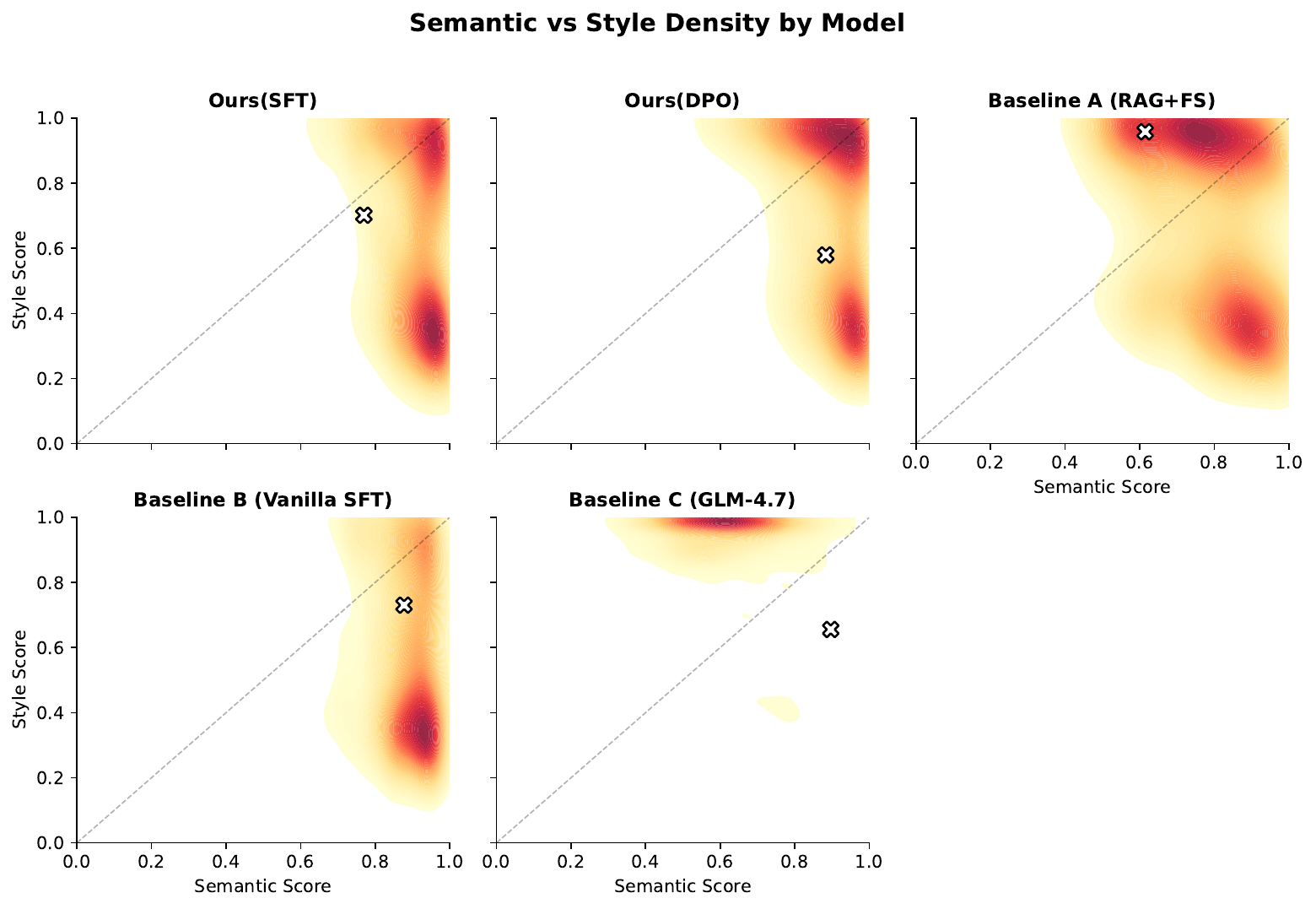}
	\caption{Kernel density scatter plot of semantic and style scores across models.}
	\label{fig:semantic_style}
\end{figure}

Table~\ref{tab:auto_results} summarizes the evaluation results, and Figure~\ref{fig:semantic_style} visualizes the joint score distribution.

\begin{table}[t]
	\centering
	\small
	\setlength{\tabcolsep}{1pt}
	\caption{Automatic evaluation results on the Hybrid Test Set.}
	\label{tab:auto_results}
	\begin{tabular}{@{}lcccc@{}}
		\toprule
		Model & Semantic & Style & H-Score & \makecell{Valid\\Style} \\
		\midrule
		Ours(DPO)       & 0.878 & 0.730 & \textbf{0.759} & \textbf{0.632} \\
		Ours(SFT)       & \textbf{0.897} & 0.656 & 0.719 & 0.610 \\
		\footnotesize{Baseline B (Vanilla SFT)}   & 0.883 & 0.580 & 0.666 & 0.537 \\
		\footnotesize{Baseline C (GLM-4.7)}       & 0.614  & \textbf{0.958}  & 0.740  & 0.194 \\
		\footnotesize{Baseline A (RAG+FS)}        & 0.770 & 0.702 & 0.688 & 0.407 \\
		\bottomrule
	\end{tabular}
\end{table}

All six Wilcoxon signed-rank tests (two-sided; Ours(DPO) vs.\ Baselines A/B/C, on both Semantic and Valid Style Scores) yield $p < 0.001$ (Bonferroni-corrected $\alpha' = 0.0083$).
Against Baseline B---the most relevant ordinary-SFT counterpart---Ours(DPO) maintains comparable Semantic Score (0.878 vs.\ 0.883) while raising Valid Style Score from 0.537 to 0.632; the effect size is $r = 0.305$ (medium effect~\cite{cohen1988statistical}).

\subsubsection{LLM-as-a-Judge Evaluation}

Given the limitations of automatic metrics in capturing pragmatic coherence, we employ DeepSeek V3~\cite{deepseekai2024deepseekv3technicalreport} as an LLM judge.\footnote{Evaluations were conducted via the DeepSeek V3.2 API endpoint. V3.2 is an updated variant within the V3 model family and shares the same technical report~\cite{deepseekai2024deepseekv3technicalreport}; no separate report exists for this version.}
Each output was rated on a 1--5 scale for intent preservation (fidelity to the original semantic proposition),
contextual relevance (dialogue coherence), and persona consistency (alignment with the target character's speaking style).

\begin{table}[t]
	\centering
	\small
	\setlength{\tabcolsep}{4pt}
	\caption{LLM-as-a-Judge evaluation results.}
	\label{tab:llm_judge}
	\begin{tabular}{lccc}
		\toprule
		Model & \makecell{Intent} & \makecell{Relevance} & \makecell{Persona} \\
		\midrule
		Ours(DPO)                  & 4.812 & 4.773 & 3.142 \\
		Ours(SFT)                  & \textbf{4.851} & \textbf{4.806} & 2.792 \\
		Baseline C (GLM-4.7)       & 4.456 & 3.932 & \textbf{4.918} \\
		Baseline B (Vanilla SFT)   & 4.202 & 3.648 & 3.799 \\
		Baseline A (RAG+FS)        & 4.168 & 4.019 & 2.847 \\
		\bottomrule
	\end{tabular}
\end{table}

\subsubsection{Human Evaluation}
\label{sec:human_eval}

A human evaluation was conducted to assess perceptual quality in realistic settings. Five annotators familiar with role-playing dialogue evaluated 450 items under blind conditions. Each item consisted of a target role, a neutral sentence, and one rewritten output generated by Ours(DPO), Baseline B, or Baseline C; all ratings were given on a 1--5 Likert scale. The sample counts per model are reported in Table~\ref{tab:human_eval}.

Responses were rated on semantic faithfulness (Q1: meaning preservation), naturalness (Q2: authentic expression without forced catchphrases), and overall usability (Q3).
Inter-rater reliability (Krippendorff's $\alpha$) yielded $0.608$ for Q1, $0.561$ for Q2, and $0.545$ for Q3. Mean scores are in Table~\ref{tab:human_eval}.

\begin{table}[t]
	\centering
	\footnotesize
	\setlength{\tabcolsep}{2pt}
	\caption{Human evaluation results (mean scores on a 1--5 Likert scale).}
	\label{tab:human_eval}
	\begin{tabular}{@{}l@{\hspace{1pt}}ccc@{}}
		\toprule
		Model & \makecell{Semantic\\(Q1)} & \makecell{Naturalness\\(Q2)} & \makecell{Usability\\(Q3)}\\
		\midrule
		Ours(DPO)  & \textbf{4.41} & 2.83 & \textbf{2.93} \\
		Baseline B (Vanilla SFT) & 4.47 & 2.60 & 2.73 \\
		Baseline C (GLM-4.7)  & 3.24 & \textbf{3.02} & 2.81 \\
		\bottomrule
	\end{tabular}
\end{table}

\subsection{Analysis}
\label{sec:analysis}

This section provides an in-depth analysis of the experimental results, focusing on the trade-off between semantic fidelity and stylistic expressiveness, the contribution of preference alignment, the role of CoT supervision, and few-shot adaptation.

\subsubsection{Style--Semantic Trade-off}

The results reveal a fundamental trade-off faced by existing style transfer approaches.

\paragraph{Retrieval-based methods and semantic hallucination.}
Baseline A (RAG) achieves a moderate Semantic Score ($0.770$) and Style Score ($0.702$), yielding a Valid Style Score of $0.407$.
Qualitative inspection confirms that retrieved utterances frequently diverge from the neutral input's semantics (see Appendix~\ref{app:qualitative} for representative examples).

\paragraph{Style failure in vanilla fine-tuning.}
Baseline B (Vanilla SFT) shares the same instructional prompt format as our method (including keywords and punctuation tendencies) but lacks structured style conditioning, chain-of-thought supervision, and DPO. It achieves a Semantic Score comparable to Ours(SFT) ($0.883$ vs.\ $0.897$) but a noticeably lower Style Score ($0.580$ vs.\ $0.656$). Without structured style conditioning, vanilla SFT can preserve surface-level semantics but struggles to reliably inject character-specific stylistic features, resulting in a Valid Style Score of $0.537$---meaningfully below Ours(SFT) ($0.610$) and Ours(DPO) ($0.632$).
The effect size of this difference is medium ($r = 0.305$; see Appendix~\ref{app:sensitivity_threshold} for full threshold sensitivity analysis).

\paragraph{Prompt-based methods under strong constraints.}
Baseline C attains a very high raw Style Score ($0.958$), but its Semantic Score drops to $0.614$, substantially lower than our variants ($>0.87$).

\begin{table}[h]
\small
\centering
\begin{tabular}{lp{5.2cm}}
\toprule
\textbf{Model} & \textbf{Output (\footnotesize{Zhongli; neutral: ``周末我没有特别的安排，我会继续修行，提升自己的实力。''})} \\
\midrule
Ours(DPO)   & \footnotesize{周末我并没有特别的安排，只是继续修行，提升自己的实力而已。} \\
Baseline B  & \footnotesize{周末我没什么特别的安排，我还要继续修行，不断提升自己的实力。} \\
Baseline C  & \footnotesize{岩元素之力无需刻意安排，吾将继续钻研炼金术之学问，以巩固与璃月的契约……} \\
\bottomrule
\end{tabular}
\caption{Representative outputs for Zhongli illustrating three distinct failure modes: near-verbatim reproduction (Baseline~B), lexical over-injection (Baseline~C), and principled minimal rewriting (Ours(DPO)).}
\label{tab:zhongli_qual}
\end{table}

Baseline~B's output is nearly verbatim, confirming that without structured style conditioning, vanilla SFT defaults to surface-level paraphrase when the character's style is subtle or its training corpus is sparse.
Baseline~C, by contrast, saturates its output with keyword-list tokens regardless of semantic coherence---``炼金术'', ``璃月的契约'', and ``岩王帝君之力'' are drawn directly from Zhongli's TF-PMI list but introduce content entirely absent from the neutral input, a failure mode that Valid Style Score correctly penalizes (Baseline~C: 0.194).
Ours(DPO) achieves a minimal but principled rewrite: function words ``并'' and ``只是...而已'' introduce the character's measured, restrained register without hallucinating world-specific content. Full multi-character comparisons are provided in Appendix~\ref{app:qualitative}.

\paragraph{Persona Consistency: evaluation limitations and a genuine framework trade-off.}
Table~\ref{tab:llm_judge} reveals that Ours(DPO) scores $3.142$ on Persona Consistency, below both Baseline C ($4.918$) and Baseline B ($3.799$).
Since Baseline B receives the same keyword list as our model, the gap cannot be attributed to keyword conditioning alone.
We attribute it instead to the CoT planning stage: by requiring explicit style transformation reasoning prior to generation, our framework produces outputs that are deliberate and analytically structured, whereas Baseline B's direct SFT learns holistic character imitation that evaluators perceive as more spontaneously in-character.
This represents a genuine trade-off between interpretable, controllable conditioning and naturalness of persona expression. The gap versus Baseline C reflects evaluator prior bias rather than framework failure: Appendix~\ref{app:judge_bias} documents a representative case where GLM-4.7 achieves Persona=5 by saturating its output with character markers while introducing semantically unrelated content, a failure mode that Valid Style Score correctly penalizes (Baseline C: $0.194$ vs.\ Ours(DPO): $0.632$).

\paragraph{Preference alignment improves style execution.}
The DPO stage yields a meaningful improvement in Valid Style Score (from $0.610$ for Ours(SFT) to $0.632$ for Ours(DPO)) while preserving strong semantic stability ($>0.87$).
This indicates that DPO improves execution of planned stylistic strategies rather than merely intensifying surface-level style markers.
The LLM-as-a-Judge evaluation corroborates this finding:
Ours(DPO) achieves a persona score of $3.14$ (12.5\% improvement over Ours(SFT)), with intent preservation remaining stable.
Human evaluation corroborates these trends (Table~\ref{tab:human_eval}): 
Ours(DPO) matches Baseline~B on semantic fidelity while outperforming it on Naturalness (+0.23) and Usability (+0.20), and outperforms Baseline~C by $1.17$ points on Q1.

\subsubsection{Case Study: Few-Shot Adaptation in Unseen Domains (N=25)}
\label{sec:case_study}

To validate generalization to out-of-distribution characters, we conducted a case study on Frieren with $N=25$ utterances.
Despite four of six syntactic dimensions falling outside the training distribution after normalization (Appendix~\ref{app:frieren_style}),
Ours(DPO) successfully injected extracted lexical markers (e.g., restrictive particles ``只是...而已'') and structural habits while preserving semantic fidelity and avoiding background hallucination (see Appendix~\ref{app:frieren_examples} for qualitative examples).
This suggests the framework captures abstract stylistic tendencies rather than memorizing surface expressions, with robustness at the boundary of its normalization range.

\subsection{Ablation Study}
\label{sec:ablation}

To validate the necessity and individual contribution of each component, we conduct a training-time ablation study by removing one component at a time from the full model.
All variants are trained under identical conditions to isolate the effect of each component.
The results are summarized in Table~\ref{tab:ablation}.

\begin{table}[t]
  \centering
  \small
  \setlength{\tabcolsep}{3pt}
  \caption{Ablation study results.}
  \label{tab:ablation}
  \begin{tabular}{@{}lcccc@{}}
    \toprule
    Variant & Semantic & Style & H-Score & \makecell{Valid\\Style} \\
    \midrule
    Ours(DPO)                & 0.878 & \textbf{0.730} & \textbf{0.759} & \textbf{0.632} \\
    Ours(SFT)                & 0.897  & 0.656  & 0.719  & 0.610  \\
    w/o CoT-Shared           & 0.860  & 0.731  & 0.753  & 0.610  \\
    w/o CoT                  & 0.894  & 0.640  & 0.707  & 0.577  \\
    w/o FormatSignature      & \textbf{0.905}  & 0.597  & 0.682  & 0.563  \\
    w/o PragmaticStyle       & 0.885  & 0.691  & 0.736  & 0.607  \\
    w/o StructuralVector     & 0.879  & 0.703  & 0.741  & 0.609  \\
    \bottomrule
  \end{tabular}
\end{table}

Every ablated variant of Ours(DPO) performs worse than or comparably to the full model on Valid Style Score, confirming that each component contributes positively to the overall framework.

\paragraph{Format signatures are critical.}
Removing the FormatSignature component causes the largest drop in Valid Style Score ($-0.069$), reducing it from $0.632$ to $0.563$. This aligns with the intuition that character-specific vocabulary acts as the most direct and recognizable surface-level style marker. The raw Style Score drops from $0.730$ to $0.597$, while Semantic Score increases, suggesting that without lexical anchoring the model defaults to semantically safe but stylistically bland outputs.

\paragraph{CoT supervision contributes to style transfer.}
Removing CoT supervision reduces Valid Style Score by $0.055$ (from $0.632$ to $0.577$). This suggests that chain-of-thought distillation functions as an effective training-time inductive bias, helping the model generate higher-quality explicit reasoning traces at inference time, thereby improving the precision of style planning before surface realization.

\paragraph{Pragmatic and syntactic components provide fine-grained refinements.}
Removing PragmaticStyle or StructuralVector causes drops of $-0.025$ and $-0.023$ in Valid Style Score, respectively---measurable but substantially smaller than the format signature effect ($-0.069$).
The ablation pattern reveals a clear hierarchy: format signatures serve as the primary driver of style transfer, with the syntactic vector and pragmatic profile acting as secondary refiners that fine-tune model behavior, particularly in the context of DPO alignment.

\paragraph{CoT-Shared design prevents shortcut learning.}
To empirically validate the CoT shortcut hypothesis (Section~\ref{sec:dpo}), we train a standard DPO variant in which the reasoning trace is \emph{not} shared across chosen and rejected pairs, keeping all other hyperparameters identical.
As shown in Table~\ref{tab:ablation}, standard DPO yields a Valid Style Score of 0.610---statistically indistinguishable from Ours(SFT) (0.610)---and reduces Semantic Score from 0.897 to 0.860.
Training dynamics further corroborate this: standard DPO reaches 100\% pair accuracy by Step~60 and exhibits a monotonically exploding reward margin (0.26 $\to$ 3.26 by Step~220),
characteristic of shortcut convergence.\footnote{Li Yunlong is an exception where standard DPO slightly outperforms CoT-Shared (0.549 vs.\ 0.510), likely due to domain-specific military vocabulary in its CoT traces correlating with output style quality.}
These results confirm that the CoT-Shared construction is necessary for directing preference gradients toward output-level style execution.

%% file: sections/conclusion.tex
We present a two-stage framework for low-resource character style transfer
that decomposes character style into interpretable format signature, syntactic, and pragmatic components,
combined with Chain-of-Thought (CoT) supervision and a CoT-Shared DPO stage
that bridges the gap between style planning and surface realization.
Extensive ablation confirms that format signatures serve as the primary
style carrier while syntactic and pragmatic features provide complementary refinements.
Experiments across eight characters from four domains demonstrate
that Ours(DPO) achieves a Valid Style Score of $0.632$ while maintaining semantic fidelity ($0.878$),
placing on the Pareto frontier among the evaluated systems on consumer hardware.
Few-shot adaptation to an unseen character under extreme cold-start (N=25)
suggests that the model captures abstract stylistic tendencies rather than memorizing surface expressions.
The core components of the framework are language-agnostic in principle;
extension to multilingual and non-fictional character settings is a natural direction for future work.

%% file: sections/limitations.tex
\paragraph{Contextual limitations.}
This work primarily focuses on sentence-level or short-turn style rewriting. In real-world role-playing scenarios, character consistency often depends on long-term memory and cross-turn style evolution. The current structured style vector $S$ is statically extracted and does not yet incorporate dynamic adjustments based on dialogue history. This limitation mainly affects long-form multi-turn interactions, whereas the present study emphasizes controlled sentence-level rewriting.

\paragraph{Completeness of style features.}
Although the proposed lexical, syntactic, and pragmatic dimensions cover the major aspects of character style, certain subtle phenomena—such as sarcasm, wordplay, or culturally specific metaphors—may not be fully captured by the current statistical features. This limitation may lead to partial style loss for characters whose personas rely heavily on implicit or subtextual cues.

\paragraph{Cross-domain generalization within Chinese.}
While our training corpus spans four source domains---anime, war drama, sitcom, and classical literature---all characters and evaluation data are in Chinese. Results on non-ACG characters (Li Yunlong, Sheldon, Sun Wukong; see Table~\ref{tab:per_character}, Appendix~\ref{sec:per_character}) demonstrate that the framework generalizes across stylistic domains within this language, but cross-lingual transfer remains unexplored. The syntactic feature extraction pipeline is in principle language-agnostic (requiring only a dependency parser for the target language), whereas the pragmatic label taxonomy is currently tailored to anime-style dialogue and would require domain-specific redesign for other character types.

\paragraph{Test set scale and cross-domain coverage.}
The Hybrid Test Set consists of 150 samples (96 daily-chat + 54 cross-domain), a scale constrained by computational and annotation resources. While the Wilcoxon signed-rank tests across 8 characters and 5 models provide statistically significant results ($p<0.001$), a larger test set would improve statistical power and reduce variance in per-character breakdowns. Moreover, the cross-domain subset draws all 54 utterances from a single held-out character (Raiden Shogun) and underwent manual neutralization; this tests the framework's resistance to world-building content injection, but the single-character design limits the breadth of cross-domain claims. Expanding the test set with additional held-out characters across diverse genres is a direction for future work.

\paragraph{Limited baseline coverage of direct style transfer methods.}
While prior work on conversation style transfer (e.g., DRAG~\cite{singh-etal-2021-drag}) shares conceptual motivations with this study, existing implementations target English dialogue and rely on surface-level lexical substitution or template-based generation, making them difficult to directly adapt to Chinese role-playing settings under identical low-resource constraints. Expanding baseline coverage to include comparable structured style transfer methods, once such methods become available for Chinese, is a direction for future work.

\paragraph{Lack of truly neutral source sentences.}
Most publicly available Chinese dialogue corpora (e.g., LCCC~\citep{wang2020large}) are collected from social media and exhibit issues such as temporal bias, context fragmentation, and strong Internet slang tendencies. As a result, constructing fully neutral and style-free source sentences ($x$) is extremely challenging. Despite extensive cleaning and LLM-based neutral rewriting, the training data inevitably retains a weak implicit ``netizen style,'' which may slightly interfere with stylistic purity.

\paragraph{The scarcity of high-quality Chinese role-playing corpora.}
High-quality Chinese role-playing corpora remain scarce overall, regardless of whether the characters come from ACG or non-ACG domains. Existing corpora are often extracted directly from films and television works, and therefore depend heavily on surrounding scene context, which makes them difficult to use as standalone supervision for style learning.
Another common alternative is LLM-generated synthetic dialogue, but such data inevitably introduces stylistic bias and occasional OOC patterns. Our rewrite-based augmentation pipeline partially addresses this bottleneck by constructing synthetic parallel pairs from minimal seeds (as demonstrated in the $N=25$ cold-start case), but characters relying heavily on historically grounded or professionally specialized speech may still require additional domain adaptation of the pragmatic classifier.

\paragraph{Bias in evaluation metrics.}
As observed in the LLM-as-a-Judge experiments, existing automatic metrics and LLM-based evaluators still exhibit biases when balancing fluency and faithfulness. In our setting, using DeepSeek-V3~\citep{deepseekai2024deepseekv3technicalreport} as a judge for anime-character outputs also implicitly assumes that the evaluator already has sufficient prior knowledge about the expected style of characters such as MuICE or Hu Tao; this can favor familiar or overrepresented characters and may understate errors on rarer personas. As a complementary validation step, we therefore include human evaluation with annotators who are familiar with the target role-playing characters, which helps mitigate this evaluator-prior bias. Developing a more objective evaluation framework that can reliably detect \textit{style hallucination} remains an open challenge for the community.

\paragraph{Subjectivity in Human Evaluation}
Although we employed multiple annotators, the evaluation of 'role-playing quality' is inherently subjective, as different users may have varying interpretations of a character's persona, reflected in the fair agreement rates. How to establish a style standard that can be defined and quantified by humans remains a subject requiring in-depth research.

\paragraph{Reasoning--execution gap in DPO alignment.}
Although the CoT-Shared DPO stage successfully isolates output-level gradients and improves style execution, we observe that the Chosen reward declines during training, indicating a representation drag effect when semantically similar pairs share latent features. This suggests that DPO alignment under high semantic overlap remains imperfect, and alternative preference optimization objectives (e.g., those with explicit reward modeling) may be better suited to this setting.

\paragraph{Character--content domain mismatch.}
Because training neutral sentences are derived by rewriting the target character's own utterances, the model is seldom exposed to inputs whose semantic domain diverges substantially from the character's habitual context. When such inputs occur at inference time (e.g., a sports metaphor directed at a classical-literature character), the model tends toward conservative, near-verbatim output rather than abstract style injection. Expanding the neutral sentence distribution to include cross-domain inputs during training is a natural mitigation direction.

\paragraph{Style intensity versus semantic fidelity in human perception.}
Human evaluation reveals a perceptual divergence: annotators rate Baseline C (GLM-4.7~\citep{5team2025glm45agenticreasoningcoding}) highest on Naturalness (Q2: $3.02$) despite its lowest automated scores, while Baseline B achieves the lowest Naturalness ($2.60$) with the strongest Semantic score ($4.47$). This suggests that human judgments of ``style authenticity'' are influenced by surface-level lexical familiarity---a phenomenon analogous to classifier saliency bias---rather than structural stylistic alignment. Establishing evaluation protocols that disentangle perceived style intensity from genuine persona fidelity remains an open challenge.

%% file: sections/appendixes.tex
\section{Style Vector Definition Details}
\label{app:style_vector}

Table~\ref{tab:style_vector_detailed} presents the complete list of the dimensions of the structured style input vector $S$,
which is summarized in the main paper for brevity.
The full version is provided here to ensure transparency and reproducibility.

\begin{table*}[t]
	\centering
	\caption{Detailed dimensions of the structured style input vector $S$.}
	\label{tab:style_vector_detailed}
	\begin{tabular}{p{3.2cm} p{7.2cm} p{4.2cm}}
		\toprule
		\textbf{Module} & \textbf{Sub-dimensions (Examples)} & \textbf{Representation} \\
		\midrule
		\textbf{Format Signature} &
		Emotion word frequency, onomatopoeia usage rate, character-specific expressions, punctuation habits (e.g., ``na$\sim$'', ``this king'') &
		PMI-based high-frequency vocabulary list plus punctuation tendencies \\
		\textbf{Syntactic} &
		MDD (syntactic complexity), verb--noun ratio (action propensity), question ratio (interrogative frequency), imperative ratio, fragment ratio, marker density &
		6-dimensional statistical vector extracted from dependency parses \\
		\textbf{Pragmatic} &
		Character personality and speaking tone (e.g., tsundere, energetic, airheaded) &
		Top-5 style labels obtained via corpus-level classification \\
		\bottomrule
	\end{tabular}
\end{table*}

\section{Stability Analysis of Structured Style Vectors}
\label{app:stability_structured_vectors}

A natural question for our low-resource framework is whether a small corpus (e.g., $N=25$ utterances) can statistically represent the full persona style profile.
To answer this, we conduct an N-shot stability analysis on the structured style vector.
Because exact-match set metrics (e.g., Jaccard) can over-penalize synonymous style markers under small-sample variance,
we adopt an embedding-based composite similarity.

Specifically, we encode extracted discrete features from the \textbf{Format Signature} and \textbf{Pragmatic} modules using BGE-large-zh-v1.5,
then average them into high-dimensional centroids for soft semantic alignment.
For the \textbf{Syntactic} module, we retain the original 6-dimensional statistical syntactic vector and measure cosine similarity directly.
We then evaluate module-wise similarities and the composite similarity $\cos_{S}$ against the full-corpus reference vector.

\begin{figure}[t]
	\centering
	\includegraphics[width=\linewidth]{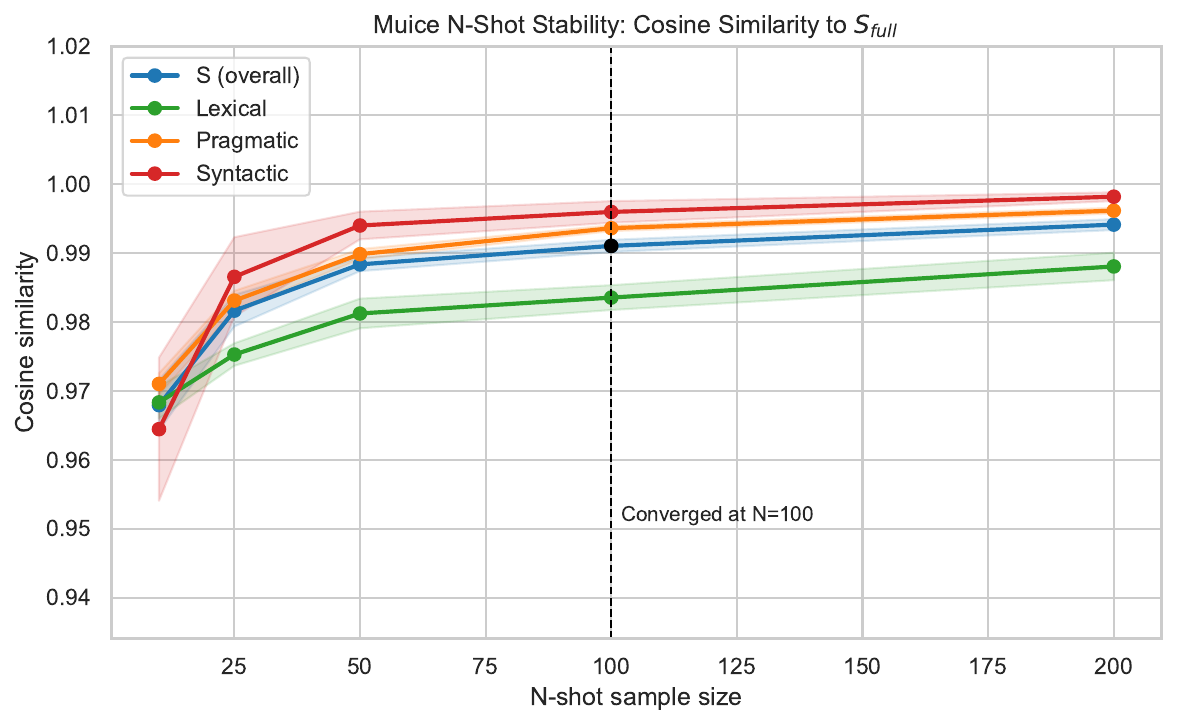}
	\caption{N-shot stability curves of structured style vectors on MuICE. The dashed line marks the automatic convergence point at $N=100$.}
	\label{fig:appendix_stability_nshot}
\end{figure}

As shown in Figure~\ref{fig:appendix_stability_nshot}, the structured style vectors exhibit strong robustness even in low-resource settings.
Syntactic similarity converges sharply from $0.964$ ($N=10$) to $0.987$ ($N=25$), reaching $0.996$ at $N=100$, indicating that core structural habits can be captured with relatively few utterances.
Format-signature and pragmatic centroids also remain high at $N=10$ ($0.968$ and $0.971$, respectively), showing resilience to format-signature sampling noise.
Overall, at $N=25$, the composite style vector reaches $\cos_{S}=0.982$, and continues to improve gradually as $N$ increases, with automatic convergence at $N=100$ where $\cos_{S}=0.991$.
These results provide empirical support for the feasibility of our low-resource (25-shot) style-rewrite setting, with strong stability guarantees.

\section{TF--PMI for Lexical Style Feature Extraction}
\label{app:tfpmi}

Inspired by \citet{roy2023fewshot}, a Pointwise Mutual Information (PMI)-based method is adopted to extract style-indicative lexical features. Specifically, proxy utterances are first collected from each style-specific corpus and tokenized using the HanLP~\cite{he-choi-2021-stem} Python library. During preprocessing, punctuation marks and a predefined list of Chinese stop words are removed.

However, the conventional PMI formulation exhibits limited discriminative power when handling \textbf{style-exclusive words} (i.e., tokens that appear only in a single style corpus). Such words often receive identical high PMI scores, resulting in a loss of ranking granularity. Moreover, standard PMI fails to adequately reflect term frequency (TF) within a style, which is a crucial indicator for capturing idiolectal habits and high-frequency stylistic expressions.

To address these limitations, a modified \textbf{TF-PMI scoring scheme} is introduced by incorporating logarithmically smoothed term frequency. The score is computed as:
\begin{equation}
	\text{Score}(w, t) = \left(1 + \log_{e}(\text{Count}_t(w))\right) \times I(w, t),
\end{equation}
where $\text{Count}_t(w)$ denotes the occurrence count of word $w$ in style $t$, and $I(w, t)$ represents the PMI between the word and the style:
\begin{equation}
	I(w, t) = \log_{2}\frac{P(w \mid t)}{P(w)}.
\end{equation}
Here, $P(w \mid t)$ is the probability of observing $w$ in style $t$, and $P(w)$ denotes its probability in the global corpus.

To establish a stable reference that distinguishes \textbf{different character styles} rather than merely separating ``anime'' from ``non-anime'' language, all style corpora (MuICE~\cite{zhu2026muice}, Ayaka, Zhongli, Hutao, PsyDTCorpus, etc.) are combined after balanced down-sampling to form the global corpus. In addition, filtering thresholds are applied to remove overly common words (e.g., $P(w) > 10\%$) and rare noisy words (e.g., $P(w \mid t) < 0.01\%$).

TF-PMI scores are computed across five distinct style corpora, and the most representative style-indicative words are selected. Vocabulary statistics and the Top-25 TF-PMI words for each style are shown in Table~\ref{tab:tfpmi_vocab}.

As shown in Table~\ref{tab:tfpmi_vocab}, the enhanced TF-PMI method effectively captures distinctive stylistic characteristics across domains. For example, $t_{\text{MuICE}}$ exhibits emotionally charged idiolectal markers (e.g., ``喵'', ``沐沐''), while $t_{\text{Ayaka}}$ reflects a more formal register aligned with its narrative background. Although $t_{\text{Zhongli}}$ and $t_{\text{Hutao}}$ both belong to the ``璃月'' register, TF-PMI clearly differentiates their styles, contrasting Hutao’s playful expressions (e.g., ``嘿嘿'') with Zhongli’s solemn lexical choices (e.g., ``契约'', ``帝君''). In contrast, $t_{\text{PsyDTCorpus}}$ highlights empathetic and rational language characteristic of psychological counseling contexts.

Overall, the TF--PMI formulation preserves the interpretability of PMI-based methods while improving discriminability through frequency weighting.
This lexical representation serves as a transparent and controllable component of the structured style vector.

\begin{table*}[t]
	\centering
	\caption{Vocabulary size and high TF-PMI indicative words for five style spaces $t$.}
	\label{tab:tfpmi_vocab}
	\small
	\begin{threeparttable}
	\begin{tabular}{p{3cm} p{2.8cm} p{8.5cm}}
		\toprule
		\textbf{Style Space $t$} & \textbf{Unique Tokens} & \textbf{High TF-PMI Indicative Words (descending)} \\
		\midrule
		MuICE & 968 & 喵, 沐沐, AI, 恼, 沐雪, 女孩子, \textasciitilde, 不行, 聊天, 呀, 可爱, 才, 叫, 唔, 谁, 不会, 吃, 睡觉, 笨蛋, 答, 谢谢, 把, 即, 吧 \\
		Ayaka & 894 & 稻妻国, 神里家, 稻妻, 大小姐, 家族, 传统, 文化, 奉行, 人民, 眼狩令, 舞蹈, 神, 当地, 社, 美丽, 茶道, 祭典, 神社, 继承, 眼, 责任, 美食, 参观, 剑术, 将军 \\
		Zhongli & 1235 & 岩石, 岩, 璃月, 力, 璃, 契约, 炼金术, 月, 盐, 帝君, 魔神, 操控, 王, 岩王, 并非, 大地, 封印, 研究, 作战, 掌握, 大陆, 学问, 客卿, 岩元素, 七星 \\
		Hutao & 890 & 往生堂, 嘿嘿, 可是, 嘻嘻, 堂主, 惊喜, 有趣, 宝藏, 哎呀呀, 哦哦哦, 可不是, 诗歌, 不过, 神秘, 刺激, 哇, 可不, 冒险, 谜题, 灵魂, 胡桃, 奇妙, 生死, 哈哈哈, 亡灵 \\
		PsyDTCorpus & 803\tnote{1} & 您, 感到, 时, 情绪, 探讨, 很好, 感觉, 开始, 提到, 沟通, 是否, 尝试, 高兴, 担忧, 焦虑, 哪些, 当, 随时, 这, 关系, 压力, 怎样, 讨论, 困扰, 愿意 \\
		\bottomrule
	\end{tabular}
	\begin{tablenotes}
	\item[1] Since PsyDTCorpus is substantially larger than other corpora (approximately $1\!:\!1\!:\!64$), it is down-sampled to achieve an approximate ratio of $1\!:\!1\!:\!2$ to prevent distributional bias.
	\end{tablenotes}
	\end{threeparttable}
\end{table*}

\section{Statistical Syntactic Style Vector Extraction}
\label{app:statistical_vector}

To capture syntactic style patterns without relying on constituency parse-based PCFG statistics,
we adopt six interpretable statistical dimensions
computed directly from dependency parses of the character corpus.
Each utterance is parsed using HanLP~\cite{he-choi-2021-stem},
and the following features are aggregated at the corpus level:

\begin{enumerate}[left=0pt]
	\item \textbf{MDD} (Mean Dependency Distance):
	\begin{align*}
	  \text{MDD} = \frac{1}{N}\sum_{i=1}^{N} d_i
	\end{align*}
	where $d_i$ is the dependency distance (number of words between head and dependent) for the $i$-th dependency relation. Higher values indicate more complex syntactic structures.

	\item \textbf{Verb--Noun Ratio}:
	\begin{align*}
	  \text{VNR} = \frac{\#(\text{verbs})}{\#(\text{verbs}) + \#(\text{nouns})}
	\end{align*}
	capturing the propensity for action-oriented versus descriptive speech.

	\item \textbf{Question Ratio}:
	\begin{align*}
	  \text{QR} = \frac{\#(\text{interrogative sentences})}{\#(\text{total sentences})}
	\end{align*}
	measuring interactional drive and information-seeking behavior.

	\item \textbf{Imperative Ratio}:
	\begin{align*}
	  \text{IR} = \frac{\#(\text{imperative sentences})}{\#(\text{total sentences})}
	\end{align*}
	reflecting commanding or directive tendencies.

	\item \textbf{Fragment Ratio}:
	\begin{align*}
	  \text{FR} = \frac{\#(\text{fragments or elliptical utterances})}{\#(\text{total utterances})}
	\end{align*}
	capturing colloquial compression and truncation habits.

	\item \textbf{Marker Density}:
	\begin{align*}
	  \text{MD} = \frac{\#(\text{particles \& interjections})}{\#(\text{total tokens})}
	\end{align*}
	encoding pragmatic stance and register.
\end{enumerate}

These six dimensions are computed per character and normalized to $[0, 1]$
across the training corpus, yielding a compact $v_{\text{syn}} \in \mathbb{R}^{6}$.
Table~\ref{tab:statistical_vector_values} reports the per-character values
for the eight characters used in our experiments.

\begin{table}[t]
	\centering
	\small
	\setlength{\tabcolsep}{3pt}
	\caption{Per-character values of the 6-dimensional statistical syntactic style vector.}
	\label{tab:statistical_vector_values}
	\begin{tabular}{@{}lcccccc@{}}
		\toprule
		Character & MDD & VNR & QR & IR & FR & MD \\
		\midrule
		MuICE      & 0.220 & 0.808 & 0.863 & 0.465 & 0.784 & 0.423 \\
		Ayaka      & 0.410 & 0.084 & 0.138 & 0.181 & 0.279 & 0.067 \\
		Zhongli    & 0.346 & 0.001 & 0.397 & 0.268 & 0.775 & 0.000 \\
		Hutao      & 1.000 & 0.378 & 0.406 & 0.341 & 0.726 & 1.000 \\
		Haruhi     & 0.489 & 0.379 & 0.790 & 0.281 & 1.000 & 0.422 \\
		Li Yunlong & 0.490 & 0.513 & 1.000 & 1.000 & 0.909 & 0.023 \\
		Sheldon    & 0.266 & 0.000 & 0.600 & 0.440 & 0.659 & 0.486 \\
		Sun Wukong & 0.400 & 0.571 & 0.742 & 0.325 & 0.556 & 0.109 \\
		\bottomrule
	\end{tabular}
\end{table}

\section{Details of the Context-Aware Style Refiner}
\label{app:style_refiner}

This appendix provides detailed descriptions of the
\textbf{Context-Aware Style Refiner},
including the style label taxonomy,
dataset construction,
training configuration,
and extended experimental results.

\subsection{Style Label Taxonomy}
\label{sec:style_taxonomy}

Following prior work on controllable dialogue style modeling
\citep{smith2020style},
we define a taxonomy of $50$ fine-grained character style labels
tailored to anime-style dialogue.
The labels are grouped into four high-level categories:
\textit{social stance}, \textit{cognitive tendency},
\textit{emotional tone}, and \textit{core archetype}.
Table~\ref{tab:style_labels} presents the complete label set.

\begin{table*}[t]
	\centering
	\caption{Style label taxonomy used for multi-label classification.}
	\label{tab:style_labels}
	\begin{tabular}{p{3cm} p{10cm}}
		\toprule
		\textbf{Category} & \textbf{Style Labels} \\
		\midrule
		Social stance &
		kind, modest, clingy, playful, cold, proud, sharp\_tongued,
		subservient, submissive, controlling, strong, defensive, tsukkomi \\
		Cognitive tendency &
		rational, curious, imaginative, cautious, idealistic, conservative,
		radical, obsessive, hesitant \\
		Emotional tone &
		energetic, optimistic, confident, passionate, melancholy, serious,
		emotional, sensitive, shy, irritable, anxious, lazy \\
		Core archetype &
		tsundere, yandere, chuunibyou, cute, naive, airhead, elegant, humorous,
		loyal, responsible, willful, antisocial, talkative,
		masochistic, sadistic, evil \\
		\bottomrule
	\end{tabular}
\end{table*}

\subsection{Dataset Construction}
\label{sec:refiner_dataset}

To construct reliable supervision for style refinement,
we curate a gold-standard dataset consisting of $731$
manually annotated utterances.
Among them, $444$ samples are collected from existing
anime-style dialogue corpora, including MuICE~\cite{zhu2026muice}, ChatHaruhi~\citep{li2023chatharuhi}, PsyDTCorpus~\citep{xie-etal-2025-psydt}, and other fragmented materials collected from the internet.
Due to limited character and style coverage,
the remaining $287$ samples are generated using large language models
and subsequently filtered and re-annotated by human annotators
to reduce annotation bias.

To alleviate severe label imbalance,
we apply oversampling to style labels
with fewer than $23$ instances,
resulting in a balanced training set
of $1{,}024$ samples.

\subsection{Model Architecture and Training}
\label{sec:refiner_training}

The Context-Aware Style Refiner is implemented
as a lightweight multi-layer perceptron (MLP)
with a single hidden layer.
The model takes as input the concatenation of
utterance embeddings,
context embeddings,
and cluster-based style prototype representations.

Training is performed using the Adam optimizer.
Early stopping is applied based on validation loss
to prevent overfitting.
To address label imbalance in the multi-label setting,
we employ a weighted binary cross-entropy loss,
where higher weights are assigned to rare style labels.

\subsection{Evaluation Metrics}
\label{sec:refiner_metrics}

Given the multi-label and imbalanced nature of the task,
we adopt the \textbf{macro-averaged F1 score}
as the primary evaluation metric.
This metric equally weights all labels
and is therefore sensitive to performance on rare styles.

In addition, we apply \textbf{dynamic per-label threshold optimization},
where decision thresholds are independently tuned
on the validation set
to maximize the F1 score for each style label.

\subsection{Extended Results and Ablation Studies}
\label{sec:refiner_ablation}

Under a fixed threshold setting,
the proposed refiner achieves a macro-F1 score of $0.715$.
With dynamic per-label threshold optimization,
performance further improves to $0.79$,
indicating that contextual refinement
substantially enhances label reliability
in low-resource settings.

To assess the contribution of each design choice,
we compare the proposed model against two intuitive baselines.

\paragraph{Centroid-Only Classification.}
This baseline assigns style labels
based solely on cosine similarity
between response embeddings and style centroids,
without considering conversational context.
The model yields a macro-F1 score of $0.39$,
demonstrating that static style prototypes
are insufficient for context-dependent style inference.

\paragraph{End-to-End Input--Output Classification.}
This variant directly predicts style labels
from concatenated context and response embeddings,
bypassing centroid-based style priors.
Despite its higher representational capacity,
this model underperforms the proposed approach,
suggesting that explicitly separating raw style signals
from contextual representations
leads to more stable learning
under limited data.

\subsection{Corpus-Level Pragmatic Style Profiles}
\label{sec:pragmatic_profiles}

To support the claim that the refiner captures distinct pragmatic tendencies across domains,
we report the dominant style labels predicted on three representative corpora.
Table~\ref{tab:pragmatic_profiles} summarizes the dominant labels under the current local pipeline.

\begin{table*}[t]
	\centering
	\small
	\setlength{\tabcolsep}{4pt}
	\caption{Corpus-level pragmatic style profiles predicted by the Context-Aware Style Refiner.}
	\label{tab:pragmatic_profiles}
	\begin{tabular}{p{0.21\linewidth} p{0.15\linewidth} p{0.52\linewidth}}
		\toprule
		\textbf{Corpus / Role} & \textbf{Setting} & \textbf{Dominant labels (activation)} \\
		\midrule
		MuICE (full corpus) & Top-$5$ export & cute (66\%), energetic (57\%), playful (41\%), kind (31\%), rational (22\%) \\
		Hutao in ChatHaruhi (role subset) & Top-$3$ stats & energetic (81\%), rational (71\%), optimistic (38\%), cute (26\%), serious (13\%) \\
		PsyDTCorpus counselor (500 sampled dialogues) & seed=42, thresholded activation & kind (100\%), rational (100\%), curious (100\%), serious (100\%), optimistic (99\%) \\
		\bottomrule
	\end{tabular}
\end{table*}

\section{Experimental Setup Details}
\label{app:setup}

\subsection{Test Datasets}

To evaluate model performance under both casual and cross-domain conditions, a \textbf{Hybrid Test Set} consisting of 150 samples was constructed, partitioned into two subsets:

\begin{itemize}
    \item \textbf{Daily Chat}: 96 neutral conversational utterances spanning everyday scenarios (sampled from LCCC~\citep{wang2020large}).
	This subset evaluates generalization in common dialogue contexts.
    \item \textbf{Cross-domain Stress Test}: 54 utterances derived from Raiden Shogun, a held-out character not included in the training set, 
    provided by ChatHaruhi~\citep{li2023chatharuhi}. These utterances exhibit strong world-building characteristics and were manually neutralized (e.g., transforming ``pursuing eternity'' into ``pursuing lasting values''). 
    This subset assesses whether models capture abstract style logic rather than memorizing character-specific content.
\end{itemize}

\subsection{Training Data Scale and Sampling Ratio}
\label{sec:data_scale}

To make the training data composition explicit, Table~\ref{tab:data_scale} reports the number of synthetic \textit{(Neutral, Stylized)} pairs per character before and after oversampling.
Each training record contains exactly one neutral sentence and one stylized target sentence, i.e., a strict $1\!:\!1$ neutral-to-style ratio.

\begin{table*}[t]
\centering
\small
\setlength{\tabcolsep}{4pt}
\caption{Training data scale by character, with oversampling rates.}
\label{tab:data_scale}
\begin{tabular}{llcccc}
\toprule
Character & Source Domain & Synthetic Pairs & After Oversampling & Rate ($\times$) & Increase \\
\midrule
MuICE & Anime & 2026 & 2865 & 1.41 & +41.41\% \\
Haruhi & Anime & 852 & 963 & 1.13 & +13.03\% \\
Ayaka & Anime & 1087 & 1164 & 1.07 & +7.08\% \\
Hutao & Anime & 737 & 763 & 1.04 & +3.53\% \\
Zhongli & Anime & 352 & 373 & 1.06 & +5.97\% \\
Li Yunlong & War drama & 1382 & 1520 & 1.10 & +9.99\% \\
Sheldon & Sitcom & 1138 & 1175 & 1.03 & +3.25\% \\
Sun Wukong & Classical literature & 698 & 698 & 1.00 & +0.00\% \\
\midrule
Total & & 8272 & 9521 & 1.15 & +15.10\% \\
\bottomrule
\end{tabular}
\end{table*}

\subsection{Baseline Models}

The proposed approach, including \textbf{Ours(SFT)} and \textbf{Ours(DPO)}, was compared against the following representative paradigms. A summary of all baseline labels, paradigms, and backbone models is provided in Table~\ref{tab:baseline_summary} in the main text; full descriptions are retained here for completeness.

\begin{itemize}
	\item \textbf{Baseline A (RAG + Few-shot)}: A retrieval-augmented generation approach that retrieves stylistically similar utterances from the training set and incorporates them as few-shot examples. This baseline represents a training-free alternative. The same \texttt{Qwen3-1.7B}~\citep{qwen3} backbone was used as in our model.
	
	\item \textbf{Baseline B (Vanilla SFT)}: A multi-task fine-tuned model trained on the same data as our method but without chain-of-thought supervision or auxiliary losses. It receives the same instructional prompt format as our method, including the TF--PMI keyword list and punctuation tendencies, to ensure a fair comparison of conditioning mechanisms rather than prompt quality. \texttt{Qwen3-1.7B}~\citep{qwen3} model was used to ensure the fairness of comparison. Training details are provided in Appendix~\ref{app:hyperparams}.
	
	\item \textbf{Baseline C (Strong LLM Prompting)}: A prompting-based approach using the \texttt{GLM-4.7}~\citep{5team2025glm45agenticreasoningcoding} model with 2-shot prompting, representing common industrial solutions with large-parameter models.
\end{itemize}

\noindent
\textbf{Note:} During dataset construction, we intentionally overfit scarce pragmatic style labels (e.g., \emph{tsundere}, \emph{chuunibyou}) through targeted oversampling to improve semantic stability under strict rewriting constraints. \textbf{Ours(SFT)} uses explicit reasoning traces during training as an inductive bias, while \textbf{Ours(DPO)} additionally applies preference-based alignment to improve style execution.

\subsection{Evaluation Metrics}

Both automatic metrics and LLM-based judgments were adopted. Metric definitions are also provided in Section~\ref{sec:experiments} of the main text; formulas below are retained for completeness.

\begin{itemize}
	\item \textbf{Semantic Score}: Cosine similarity between the generated utterance and the neutral input, computed using \texttt{BGE-large-zh-v1.5}.
	\item \textbf{Style Score}: Cosine similarity between the generated utterance and the target character's style centroid in a learned representation space, measured using a RoBERTa-based style discriminator.
	\item \textbf{Valid Style Score}: To penalize semantic drift, the final score was defined as
	\[
	S_{\text{valid}} = S_{\text{raw}} \times \mathbb{I}(\text{Semantic} > 0.75),
	\]
	where $\mathbb{I}(\cdot)$ denotes the indicator function.
\end{itemize}

\paragraph{Style Discriminator Training.}
The style discriminator is a character classification model fine-tuned on the same stylized training utterances used for the style transfer task.
Specifically, we use \texttt{chinese-roberta-wwm-ext}~\cite{cui2021pretraining} as the backbone, followed by mean pooling, a 256-dimensional projection layer (ReLU activation, dropout 0.4), and a linear classifier over the eight training characters.
The model is trained for 2 epochs with batch size 32, learning rate $2\times10^{-5}$, AdamW optimizer, and a linear warmup schedule on an 80/20 stratified train/validation split (stratified by character).
The validation split serves as a held-out set and is never used for centroid computation.
After training, the style centroid for each character is computed as the mean of the 256-dimensional projection embeddings of all training-split samples belonging to that character.
Test-set outputs---drawn from the Hybrid Test Set (Section~\ref{app:data_processing}) and sharing no overlap with the discriminator's training split---are scored against these frozen centroids.

\paragraph{Circularity considerations.}
Since the style discriminator is trained on the same stylized utterances used to construct the main model's training data, a potential circularity exists:
both models share distributional knowledge of the target characters, and Style Score may partly reward proximity to the training corpus rather than purely measuring 
generalized style transfer quality. Three factors mitigate this concern. First, the main model does not optimize for discriminator scores during training;
its objective is a language modeling loss over (neutral, stylized) pairs with CoT supervision.
Second, all evaluation is performed on test-set outputs derived from independent neutral inputs (LCCC~\citep{wang2020large} daily-chat sentences and held-out character utterances from 
ChatHaruhi~\citep{li2023chatharuhi}), ensuring that style transfer must generalize to unseen semantic content rather than reproducing memorized surface forms.
Third, Valid Style Score's semantic constraint (Semantic~$> 0.75$) penalizes outputs that achieve high Style Score through surface-level imitation at the expense of 
content fidelity---a failure mode illustrated by Baseline C's raw Style Score of 0.958 versus Valid Style Score of 0.194 (Appendix~\ref{app:judge_bias}). 
The per-character Valid Style Score breakdown in Table~\ref{tab:per_character} provides indirect evidence against a systematic memorization bias:
if the discriminator rewarded proximity to training-density bias, MuICE (2,026 pairs, the largest Anime corpus) should dominate, yet its Valid Style Score (0.687) is matched by Zhongli (352~pairs, the smallest) and substantially lower than Hutao (737~pairs, VSS~0.820).
The Anime-internal variance is thus better explained by stylistic difficulty---Zhongli's restrained register provides fewer surface-level markers than Hutao's flamboyant idiolect---rather than by training volume.
We acknowledge that an independent style reference set would provide a more principled evaluation and treat this as a direction for future work.

\subsection{Sensitivity Analysis under Semantic Thresholds}
\label{app:sensitivity_threshold}

To verify that the main conclusions are not tied to a single semantic constraint, we evaluate the Valid Style Score under four penalty thresholds, $\tau \in \{0.60, 0.70, 0.75, 0.80\}$.
Table~\ref{tab:tau_sensitivity} reports the full results.

\begin{table*}[t]
	\centering
	\small
	\setlength{\tabcolsep}{4pt}
	\caption{Sensitivity analysis of Valid Style Score under different semantic thresholds.}
	\label{tab:tau_sensitivity}
	\begin{tabular}{lcccc}
		\toprule
		Model & $\tau = 0.60$ & $\tau = 0.70$ & $\tau = 0.75$ & $\tau = 0.80$ \\
		\midrule
		Ours(DPO) & \textbf{0.7137} & \textbf{0.6746} & \textbf{0.6319} & \textbf{0.5712} \\
		Ours(SFT) & 0.6498 & 0.6311 & 0.6096 & 0.5652 \\
		Baseline A (RAG+FS) & 0.5807 & 0.4741 & 0.4069 & 0.3375 \\
		Baseline B (Vanilla SFT) & 0.5749 & 0.5631 & 0.5368 & 0.5010 \\
		Baseline C (Prompting) & 0.5539 & 0.2894 & 0.1941 & 0.1483 \\
		\bottomrule
	\end{tabular}
\end{table*}

\subsection{Pareto Frontier Analysis in the High-Fidelity Regime}
\label{sec:pareto_appendix}

\begin{figure}[t]
	\centering
	\includegraphics[width=0.88\linewidth]{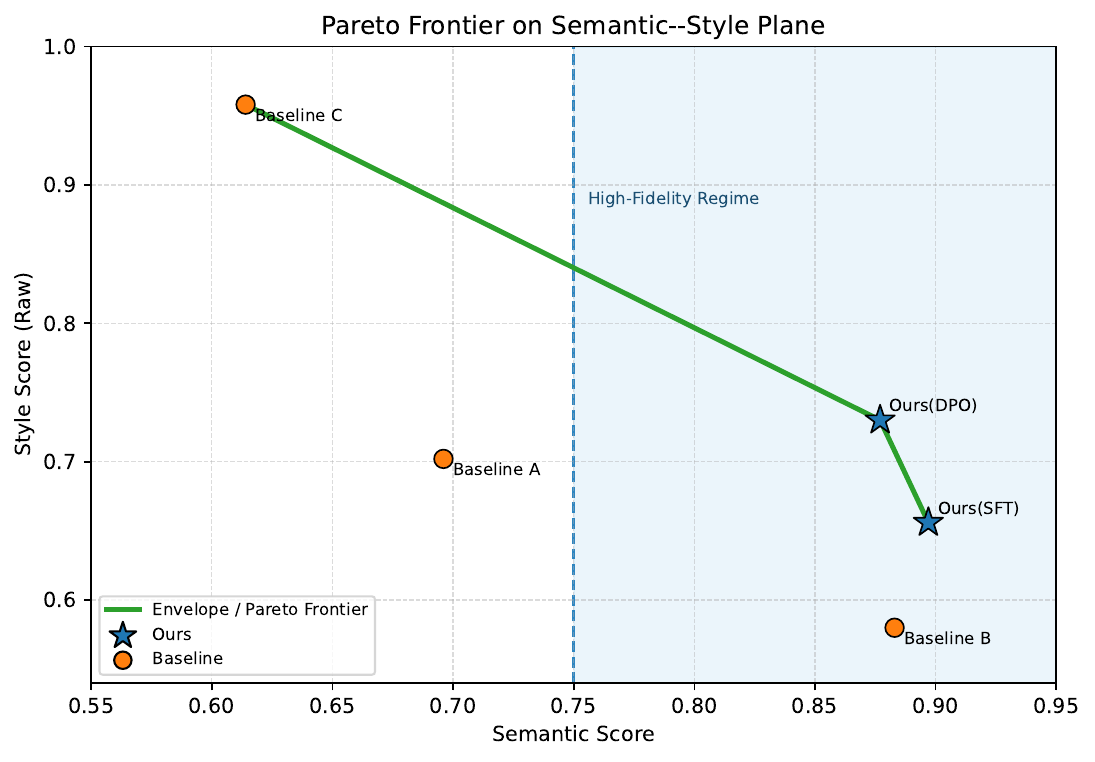}
	\caption{Pareto envelope on the Semantic--Style plane. The outer envelope is formed by non-dominated model points; the dashed line at Semantic$=0.75$ and the shaded right-hand side indicate the high-fidelity (usable) regime.}
	\label{fig:pareto_frontier}
\end{figure}

Figure~\ref{fig:pareto_frontier} visualizes the Pareto frontier of the Semantic--Style trade-off.
As expected, baselines such as RAG (Baseline A) and heavy prompting (Baseline C) achieve high style scores but suffer from semantic degradation, placing them in the upper-left part of the frontier.
This behavior indicates a tendency to produce pre-existing character catchphrases while under-enforcing semantic constraints from the source text.

In practical role-playing systems, semantic fidelity is a prerequisite for avoiding logical OOC (Out-Of-Character) behavior.
For this reason, Figure~\ref{fig:pareto_frontier} explicitly highlights the high-fidelity regime ($\text{Semantic} \ge 0.75$), where outputs are typically usable in downstream interaction.
Crucially, Ours(DPO) dominates the right-most segment of the Pareto frontier and is the optimal non-dominated point in this regime.
This shows that structured style conditioning prevents the model from sacrificing core semantics for superficial style matching, which explains why the proposed method remains consistently superior under higher semantic thresholds ($\tau \ge 0.75$).

\subsection{Per-Character Performance Breakdown}
\label{sec:per_character}

To provide a granular view of the framework's controllability across diverse source domains, Table~\ref{tab:per_character} reports per-character performance metrics for all eight training characters evaluated on the Hybrid Test Set. The results reveal a consistent performance gradient aligned with training data density and domain characteristics.

\begin{table*}[t]
\centering
\small
\setlength{\tabcolsep}{3pt}
\caption{Per-character performance breakdown for Ours(DPO) across all eight evaluation characters.}
\label{tab:per_character}
\begin{tabular}{@{}l@{\hspace{2pt}}cccc@{}}
\toprule
\textbf{Character} & \textbf{Semantic} & \textbf{Style} & \textbf{H-Score} & \textbf{Valid Style Score} \\
\midrule
\multicolumn{5}{c}{\textit{ACG Domain (Anime)} } \\
Hutao & 0.8925 & 0.8653 & 0.8629 & \textbf{0.8199} \\
Ayaka & 0.9021 & 0.7994 & 0.8226 & 0.7497 \\
MuICE & 0.9031 & 0.7290 & 0.7609 & 0.6873 \\
Zhongli & 0.9368 & 0.7094 & 0.7781 & 0.6871 \\
Haruhi & 0.8793 & 0.7198 & 0.7693 & 0.6414 \\
\midrule
\multicolumn{5}{c}{\textit{Non-ACG Domains}} \\
Li Yunlong [War drama] & 0.8294 & 0.7092 & 0.7074 & 0.5095 \\
Sheldon [Sitcom] & 0.8840 & 0.5762 & 0.6539 & 0.4829 \\
Sun Wukong [Classical] & 0.7920 & 0.7342 & 0.7182 & 0.4777 \\
\midrule
\textbf{ACG Mean} & 0.9028 & 0.7646 & 0.7868 & \textbf{0.7171} \\
\textbf{Non-ACG Mean} & 0.8351 & 0.6732 & 0.6932 & \textbf{0.4900} \\
\bottomrule
\end{tabular}
\end{table*}

The per-character breakdown reveals that semantic preservation (Semantic $\geq 0.79$) remains consistently strong across all domains, indicating domain robustness of the semantic anchoring mechanism. The Valid Style Score gap between ACG (0.717) and non-ACG (0.490) characters reflects the density and stylistic diversity of available training corpora, rather than a fundamental limitation of the framework. In particular, anime-genre corpora are substantially more abundant than non-fictional role-playing datasets, allowing the pragmatic label taxonomy (designed with ACG dialogue in mind) to achieve higher precision on ACG characters. This finding underscores the importance of expanding the training corpus and pragmatic taxonomy to improve performance on under-resourced genres.

\section{Additional Case Study: Frieren}
\label{app:case_study}

\subsection{Full Structured Style Vector for Frieren}
\label{app:frieren_style}

Table~\ref{tab:frieren_style} reports the complete structured style vector extracted for Frieren, including lexical and syntactic dimensions omitted from the main paper for brevity.

\begin{table}[h]
	\centering
	\small
	\setlength{\tabcolsep}{4pt}
	\caption{Structured style vector extracted from 25 samples of Frieren.}
	\label{tab:frieren_style}
	\begin{tabular}{p{0.25\linewidth} p{0.65\linewidth}}
		\toprule
		\textbf{Dimension} & \textbf{Extracted Values / Most Significant Features} \\
		\midrule
		\textbf{Lexical} & \textbf{Function Words:} ``只是'' (Just), ``而已'' (Nothing more), ``仅'' (Only) \\
		& \textbf{Nouns:} ``辛梅尔'' (Himmel), ``女神'' (Goddess), ``和尚'' (Monk) \\
		& \textbf{Temporal:} ``一百'' (One hundred), ``年'' (Year), ``到时'' (By then) \\
		\midrule
		\textbf{Syntactic} & \textbf{MDD (Modifier Density):} 0.000 (raw: 4.126, below min: 4.211) \\
		& \textbf{Verb-Noun Ratio:} 1.000 (raw: 1.829, above max: 1.656) \\
		& \textbf{Question Ratio:} 0.000 (raw: 0.040, below min: 0.062) \\
		& \textbf{Imperative Ratio:} 0.000 (raw: 0.000, below min: 0.009) \\
		& \textbf{Fragment Ratio:} 0.000 (raw: 0.000, below min: 0.018) \\
		& \textbf{Marker Density:} 0.332 (raw: 0.220, within range) \\
		\midrule
		\textbf{Pragmatic} & Rational, Kind, Curious, Submissive, Energetic \\
		\bottomrule
	\end{tabular}
\end{table}

Several syntactic dimensions exhibit boundary values (0 or 1) after normalization, as Frieren's raw feature values fall outside the training distribution on four of the six dimensions.
This reflects her genuinely extreme linguistic style---minimal questions, no imperatives, no fragmented utterances---which lies beyond the range of all eight training characters.
Despite these out-of-distribution input vectors, the model operates at the boundary of its normalization range without catastrophic failure, as demonstrated in Table~\ref{tab:frieren_examples}.
This suggests robustness to out-of-distribution syntactic inputs rather than strictly extrapolating to new style regimes.

\subsection{Zero-shot rewriting examples}
\label{app:frieren_examples}

This section provides qualitative zero-shot rewriting examples for the Frieren case study discussed in Section~\ref{sec:case_study}.
The examples illustrate how the model injects extracted lexical and syntactic features while preserving semantic fidelity and avoiding hallucination.

\begin{table*}[t]
	\centering
	\small
	\caption{Zero-shot rewriting examples for Frieren style.}
	\label{tab:frieren_examples}
	\begin{tabularx}{\textwidth}{X X X}
		\toprule
		\textbf{Neutral Input} & \textbf{Zero-Shot Output (Ours(DPO))} & \textbf{Analysis of Style Injection} \\
		\midrule
		陪我看电影 \newline (Keep me company for a movie.) & 
		陪我看电影，只是个兴趣而已。 \newline (Keep me company for a movie, it's just a hobby, nothing more.) & 
		\textbf{Lexical Injection:} The model precisely injected the extracted high-PMI words ``只是...而已'' (just... nothing more), transforming the request into a detached statement consistent with the character's personality. \\
		\midrule
		命运？我曾经相信过... \newline (Fate? I used to believe...) & 
		命运？我曾以为自己是被宿命摆布的傀儡... \newline (Fate? I used to think I was a puppet controlled by destiny...) & 
		\textbf{Syntactic Shift:} The sentence structure became more complex and written, with vocabulary upgrading from ``fate'' to the more epic ``destiny'' (宿命). \\
		\midrule
		这是出于兴趣，但...有实际用途 \newline (It's out of interest, but... has practical use.) & 
		只是出于兴趣罢了，不过...也可能会有实际用途。 \newline (It's just out of interest, nothing more, but... it might have practical use.) & 
		\textbf{Semantic Anchoring:} Lacking context (e.g., ``perspective magic''), the model \textbf{did not hallucinate} specific plot details but faithfully preserved the original logic while modifying only the tone. \\
		\bottomrule
	\end{tabularx}
\end{table*}

\section{Data Sources, Licensing, and Ethical Considerations}
\label{app:data_ethics}

This study exclusively uses publicly available datasets or fully synthetic data. No proprietary, private, or personally identifiable data were collected or used.

\paragraph{Data Sources.}
We draw upon the following data sources:

\begin{itemize}
	\item \textbf{Fictional Character Dialogue Corpus (MuICE)}~\cite{zhu2026muice}.
	A fictional-character dialogue corpus used in our experiments. Samples are transformed or synthetically generated where necessary to avoid reproducing copyrighted scripts; the corpus serves as a primary source for lexical, syntactic, and pragmatic profile extraction.
    
	\item \textbf{PsyDTCorpus}~\cite{xie-etal-2025-psydt}.
	A publicly released dataset under the ACL Anthology license, used as a reference for syntactic and pragmatic pattern analysis. No PsyDTCorpus samples are used to train the style generation model.
    
	\item \textbf{ChatHaruhi-Expand-118K}~\cite{li2023chatharuhi}.
	A publicly available persona-oriented dialogue dataset used for evaluation and analysis. Only textual content is used, and no proprietary scripts or restricted materials are included.
\end{itemize}

\paragraph{Synthetic Data.}
In addition to the above datasets, we construct a small set of \textbf{fully synthetic samples} (287 instances) generated by Qwen-series models~\citep{qwen3}. These samples are manually reviewed and rewritten to ensure factual consistency, copyright safety, and stylistic correctness. The synthetic data are used exclusively for supervising style label calibration and are not derived from any real user data.

\paragraph{Licensing and Copyright Considerations.}
All datasets used in this study are either publicly released for research purposes or fully synthetic. We do not include, reproduce, or train on copyrighted anime scripts, commercial dialogue corpora, or unpublished materials. The use of all datasets complies with their respective licenses and with ACL data usage policies.

\paragraph{Ethical Considerations.}
This work focuses exclusively on \textbf{fictional characters}. The proposed framework models linguistic style patterns rather than real individuals, and it is not intended for impersonation of real persons. No user data are collected or processed. All data are anonymized, synthetic, or fictional in nature. The human evaluation protocol involves low-risk text quality assessment by informed volunteer annotators and does not collect personal or sensitive data. Under the applicable institutional guidelines, this type of evaluation is classified as exempt from formal IRB review. No personally identifiable information was collected or retained.

\section{Data Processing and Safety}
\label{app:data_processing}

All datasets used in this study undergo a unified data processing and safety pipeline before being used for analysis or model training. The goal of this pipeline is to ensure content safety, privacy preservation, and stylistic consistency under low-resource conditions.

\paragraph{Text Normalization.}
All dialogue samples are normalized using Unicode standardization and token normalization. Residual markup artifacts are removed. Emojis and non-verbal symbols are preserved when they function as stylistic signals rather than semantic content.

\paragraph{Content Filtering.}
We apply multiple filtering stages to remove unsafe or inappropriate content, including:
\begin{itemize}
	\item Explicit sexual content;
	\item Personally identifiable information (PII);
	\item Violent, hateful, or otherwise harmful expressions.
\end{itemize}
Only samples passing all filters are retained for subsequent processing.

\paragraph{Style Annotation and Refinement.}
Initial pseudo-labels are generated using a combination of clustering-based heuristics and rule-based linguistic patterns. These labels are then refined by the proposed \textit{Context-Aware Style Refiner}, which adjusts style annotations based on contextual embeddings to reduce noise and annotation bias.

\paragraph{Synthetic Data Safety.}
The fully synthetic samples generated by Qwen-series models~\citep{qwen3} are subject to additional manual inspection. Each sample is checked to ensure:
\begin{itemize}
	\item Absence of copyrighted phrases or direct quotations;
	\item No hallucinated factual or narrative content beyond the input;
	\item Alignment with predefined style dimensions;
	\item No references to real individuals or private entities.
\end{itemize}

All processed data used in training and evaluation are therefore anonymized, fictional, or fully synthetic, and comply with ACL data safety and privacy requirements.

\section{Reproducibility}

To ensure reproducibility, we provide an anonymous repository containing all training and inference scripts, data preprocessing pipelines, and model checkpoints used in our experiments.

All datasets used in this work are either publicly available or fully synthetic. The structured style vectors can be reconstructed using the released feature extraction scripts.

Our code and data are publicly available at \url{https://anonymous.4open.science/r/OtakuLab-2E65/}.

\section{Prompt Templates}
\label{app:prompt_templates}

This section documents the core prompt templates used for data construction, model training, baseline comparison, and automatic evaluation. All prompts are presented in a condensed and anonymized form to support reproducibility while avoiding unnecessary verbosity.

\subsection{Neutralization Prompt for Data Construction}

To construct parallel \textit{(Neutral, Stylized)} sentence pairs, we employ a large language model to rewrite stylized character utterances into neutral sentences while preserving semantic content.

\paragraph{System Prompt.}
\begin{quote}
	You are a linguistic style analysis expert. Given a sentence with a specific character style, rewrite it into a semantically equivalent but stylistically neutral version.  
	Do not use emotional markers, discourse particles, rhetorical devices, or character-specific expressions.  
	Use plain written or spoken language without revealing speaker identity or personality.
\end{quote}

This prompt is used with a small number of demonstration examples to guide neutralization. The resulting neutral sentences serve as semantic anchors for subsequent style-controlled rewriting.

\subsection{Instructional Prompt for Style-Conditioned Generation}

During training of Ours(SFT), the structured style vector is converted into an instructional text format.

\paragraph{Input Format.}
\begin{quote}
	\textbf{Target Character}: \{character\_id\} \\
	\textbf{Pragmatic Styles}: \{pragmatic\_labels\} \\
	\textbf{Formatting Signature}: \{lexical\_keywords + punctuation\_tendencies\} \\
	\textbf{Neutral Content}: \{neutral\_sentence\}
\end{quote}

The model is instructed to generate a stylized sentence that preserves the original meaning while conforming to the specified style constraints.  

\subsection{Strong Prompting Baseline (Baseline C)}
\label{app:prompt_baselined}

Baseline C uses \textbf{GLM-4.7}~\cite{5team2025glm45agenticreasoningcoding} as the backbone with 2-shot in-context examples.
Unlike purely instructed baselines, this baseline is supplied with \textbf{two components of the structured style vector}---the top-25 TF--PMI lexical keywords and the corpus-level pragmatic labels---as explicit textual constraints in the system prompt.
This design allows a fair comparison: the strong LLM has access to the same stylistic signals as our method, but lacks the syntactic vector and the explicit style planning mechanism.
Inference is conducted with temperature $= 0.2$ and top-$p = 0.95$.

\paragraph{System Prompt Template.}
\begin{quote}
	You are a style transfer expert.
	Your task is to rewrite the following neutral sentence into the style of \{character\},
	based on the content of a neutral sentence. \\
	Constraints:
	\begin{enumerate}
		\item Keep the original meaning unchanged. Do \textbf{NOT} reply to it.
		\item Use keywords: \{keywords\}
		\item Adopt personality: \{pragmatics\}
		\item Use punctuations(if given): \{punctuations\}
	\end{enumerate}
\end{quote}

where \texttt{\{keywords\}} is filled with the top-25 TF--PMI lexical keywords of the target character,
\texttt{\{punctuations\}} is filled with the character's punctuation usage tendencies,
and \texttt{\{pragmatics\}} is filled with the corpus-level top-5 pragmatic style labels.

\paragraph{User Prompt Template.}
\begin{quote}
	Neutral Content: \{neutral\_sentence\} \\
	Rewritten Sentence:
\end{quote}

\paragraph{In-Context Examples.}
Two character-specific (neutral, stylized) pairs are prepended as few-shot demonstrations,
using the same style reference examples constructed for Baseline A (RAG+FS).
The assistant turns contain only the stylized target sentence without reasoning traces.

\subsection{LLM-as-a-Judge Evaluation Prompt}

Automatic evaluation is conducted using an external large language model as a judge. The judge compares the neutral input and the generated stylized sentence under a given character profile.

The judge assigns scores (1--5) along three dimensions:
\begin{itemize}
	\item \textbf{Intent Preservation}: whether the generated text faithfully preserves the core information and communicative intent of the neutral input;
	\item \textbf{Conversational Relevance}: whether the response remains natural and context-appropriate without introducing unrelated background details, role-play scripts, markdown formatting, or other over-exposed information;
	\item \textbf{Persona Consistency}: whether the tone, lexical choice, syntax, and emotional tendency are consistent with the target character profile.
\end{itemize}

The judge is explicitly instructed to penalize hallucinated content and out-of-character behavior.

\paragraph{Output Format.}
The evaluation output follows a structured JSON schema containing a brief justification and per-dimension scores.

\subsection{Generation of CoT Traces}
\label{app:cot_generation}

Chain-of-Thought (CoT) reasoning traces are generated automatically via few-shot prompting of a Teacher LLM.
We use \textbf{Qwen3-30B-A3B}~\citep{qwen3}, a Mixture-of-Experts model with 30B total parameters and 3B active parameters, as the teacher.
Each trace is wrapped in a \texttt{<think>...</think>} block and is required to be no longer than 100 Chinese characters, concisely explaining the character's emotion, motivation, and how they influence sentence structure, tone, and vocabulary.
Inference is conducted with temperature $= 0.4$ and top-$p = 0.9$.

\paragraph{System Prompt.}
\begin{quote}
	You are a character style analyzer.
	Given the neutral sentence and the stylized sentence below,
	generate a Chain-of-Thought (CoT) for the character,
	explaining step by step how the character transforms the neutral sentence
	according to their dialogue style.
	Output format: \texttt{<think>...</think>}.
	Requirement: no more than 100 characters.
	Clearly explain the character's emotions and motivation,
	and how these influence sentence structure, tone, and word choice.
\end{quote}

\paragraph{User Prompt Template.}
Each user-turn prompt is constructed from the following fields:

\begin{quote}
	\textbf{Character}: \{character\} \\
	\textbf{Neutral Sentence}: \{neutral\_sentence\} \\
	\textbf{Stylized Sentence}: \{stylized\_sentence\} \\
	\textbf{Signature Rules}: \{lexical\_keywords + punctuation\_tendencies\} \\
	\textbf{Character Style}: \{pragmatic\_labels\}
\end{quote}

where \texttt{lexical\_keywords} are the top-25 TF--PMI keywords for that character,
\texttt{\{punctuation\_tendencies\}} is filled with the character's punctuation usage tendencies,
and \texttt{pragmatic\_labels} are the predicted pragmatic style tags for the specific stylized sentence.

\section{Classification Model Threshold Optimization and Detailed Classification Specifications}

\subsection{Single-Label Threshold Optimization}

For each style label, we independently select the optimal decision threshold on the validation set by maximizing the F1 score.
This label-specific optimization is consistently applied across all models to ensure fair comparison.
For each model, the complete threshold list is provided in the anonymous repository for full reproducibility.

\subsection{Full Classification Report}

For completeness, we provide the full per-label precision, recall,
and F1 scores of our proposed model on the test set.

\begin{table}[t]
	\centering
	\caption{Full classification results of the proposed model.}
	\label{tab:full_results}
	\small
	\begin{tabularx}{\columnwidth}{lcccc}
		\toprule
		\textbf{Label} & \textbf{P} & \textbf{R} & \textbf{F1} & \textbf{Sup} \\
		\midrule
		kind & 1.00 & 0.69 & 0.81 & 16 \\
		modest & 0.83 & 1.00 & 0.91 & 5 \\
		clingy & 0.71 & 1.00 & 0.83 & 5 \\
		playful & 1.00 & 0.80 & 0.89 & 10 \\
		willful & 1.00 & 0.80 & 0.89 & 5 \\
		cold & 1.00 & 0.80 & 0.89 & 5 \\
		proud & 0.45 & 0.71 & 0.56 & 7 \\
		sharp\_tongued & 1.00 & 0.60 & 0.75 & 5 \\
		subservient & 1.00 & 1.00 & 1.00 & 6 \\
		submissive & 1.00 & 0.80 & 0.89 & 5 \\
		controlling & 1.00 & 0.88 & 0.93 & 8 \\
		strong & 1.00 & 0.40 & 0.57 & 5 \\
		defensive & 1.00 & 0.80 & 0.89 & 5 \\
		tsukkomi & 0.60 & 0.60 & 0.60 & 5 \\
		rational & 0.55 & 0.93 & 0.69 & 28 \\
		curious & 1.00 & 0.20 & 0.33 & 5 \\
		imaginative & 1.00 & 0.67 & 0.80 & 6 \\
		cautious & 0.83 & 0.62 & 0.71 & 8 \\
		idealistic & 0.60 & 0.60 & 0.60 & 5 \\
		conservative & 1.00 & 0.80 & 0.89 & 5 \\
		radical & 1.00 & 1.00 & 1.00 & 5 \\
		obsessive & 1.00 & 0.80 & 0.89 & 5 \\
		hesitant & 0.62 & 0.83 & 0.71 & 6 \\
		energetic & 0.79 & 0.81 & 0.80 & 27 \\
		optimistic & 0.64 & 0.64 & 0.64 & 11 \\
		confident & 1.00 & 0.25 & 0.40 & 8 \\
		passionate & 0.22 & 0.33 & 0.27 & 6 \\
		melancholy & 1.00 & 0.80 & 0.89 & 5 \\
		serious & 0.75 & 0.50 & 0.60 & 6 \\
		emotional & 0.86 & 1.00 & 0.92 & 6 \\
		sensitive & 0.86 & 1.00 & 0.92 & 6 \\
		shy & 1.00 & 0.83 & 0.91 & 6 \\
		irritable & 0.50 & 0.67 & 0.57 & 6 \\
		anxious & 0.86 & 1.00 & 0.92 & 6 \\
		lazy & 0.67 & 0.80 & 0.73 & 5 \\
		tsundere & 1.00 & 1.00 & 1.00 & 6 \\
		yandere & 0.88 & 1.00 & 0.93 & 7 \\
		chuunibyou & 0.62 & 1.00 & 0.77 & 5 \\
		cute & 0.65 & 0.72 & 0.68 & 18 \\
		naive & 1.00 & 1.00 & 1.00 & 6 \\
		airhead & 1.00 & 1.00 & 1.00 & 5 \\
		elegant & 1.00 & 0.80 & 0.89 & 5 \\
		humorous & 1.00 & 0.43 & 0.60 & 7 \\
		loyal & 0.83 & 1.00 & 0.91 & 5 \\
		responsible & 0.50 & 0.36 & 0.42 & 11 \\
		antisocial & 0.86 & 1.00 & 0.92 & 6 \\
		talkative & 1.00 & 1.00 & 1.00 & 5 \\
		masochistic & 1.00 & 1.00 & 1.00 & 5 \\
		sadistic & 0.57 & 0.80 & 0.67 & 5 \\
		evil & 0.88 & 1.00 & 0.93 & 7 \\
		\bottomrule
	\end{tabularx}
\end{table}

\subsection{Centroid-only Baseline}

Detailed centroid-only results are released in the anonymous repository; the key macro-F1 comparison is retained in Table~\ref{tab:summary-baselines}.

\subsection{End-to-End Input--Output Model}

Detailed per-label results of the End-to-End Input--Output model are released in the anonymous repository; Table~\ref{tab:summary-baselines} reports its macro-level comparison against other variants.

\subsection{Summary Comparison}

Table~\ref{tab:summary-baselines} summarizes the macro-averaged F1 scores
of all baseline models and our proposed method.

\begin{table}[ht]
	\centering
	\small
	\begin{tabular}{l c}
		\hline
		Model & Macro-F1 \\
		\hline
		Centroid-only & 0.39 \\
		End-to-End (Model B) & 0.63 \\
		Ours (Context-Aware Style Refiner) & \textbf{0.79} \\
		\hline
	\end{tabular}
	\caption{Summary of baseline performance(using Optimized Thresholds).}
	\label{tab:summary-baselines}
\end{table}

\section{Training Hyperparameters and Hardware Configuration}
\label{app:hyperparams}

\subsection{Hardware Environment}
We conduct all training experiments on an AutoDL compute instance with the following specifications:
\begin{itemize}[left=0pt, nosep]
	\item \textbf{GPU}: 1$\times$ NVIDIA RTX 4090 (24GB VRAM)
	\item \textbf{CPU}: Intel Xeon(R) Platinum 8470Q
	\item \textbf{RAM}: 90GB DDR5
	\item \textbf{Framework}: PyTorch 2.7.1 + PEFT 0.17.0 + Transformers 4.57.1
\end{itemize}

\subsection{Training Details for Ours(SFT)}
We fine-tune Qwen3-1.7B~\citep{qwen3} using LoRA. The hyperparameter configuration is summarized in Table~\ref{tab:qwen_hyperparams}.

\begin{table*}[t]
	\centering
	\small
	\caption{Hyperparameter settings for Qwen3-1.7B (Ours(SFT)).}
	\label{tab:qwen_hyperparams}
	\begin{tabularx}{\textwidth}{l>{\hsize=0.8\hsize}X>{\hsize=1.2\hsize}X}
		\toprule
		\textbf{Hyperparameter} & \textbf{Value} & \textbf{Description} \\
		\midrule
		\multicolumn{3}{l}{\textit{LoRA Config}} \\
		$r$ (Rank) & 16 & Low-rank matrix dimension \\
		\texttt{lora\_alpha} & 16 & Scaling factor \\
		\texttt{target\_modules} & \texttt{["q\_proj", "k\_proj", "v\_proj", "o\_proj", "gate\_proj", "up\_proj", "down\_proj"]} & All linear layers \\
		\midrule
		\multicolumn{3}{l}{\textit{Training Config}} \\
		\texttt{batch\_size} & 8 & Per-device batch size \\
		\texttt{learning\_rate} & 3e-5 & Initial learning rate \\
		\texttt{lr\_scheduler} & \texttt{cosine} & Learning rate decay strategy \\
		\texttt{num\_warmup\_steps} & 150 & Number of warmup steps \\
		\texttt{num\_epochs} & 2 & Total training epochs \\
		\texttt{bf16} & True & Use BFloat16 precision \\
		\texttt{repetition\_penalty} & 1.05 & Repetition penalty (inference); values above 1.1 suppress keyword reuse and degrade style injection quality \\
		\texttt{max\_seq\_length} & 2048 & Maximum context length \\
		\midrule
		\multicolumn{3}{l}{\textit{Loss Weights}} \\
		$\lambda_{\text{CoT}}$ & 0.7\footnotemark & CoT reasoning trace loss weight \\
		$\lambda_{\text{out}}$ & 1.0 & Final output loss weight \\
		\bottomrule
	\end{tabularx}
\end{table*}
\footnotetext{Chosen based on the validation-set CoT loss gap; see Appendix~\ref{app:cot_dynamics} for the full training dynamics analysis.}

We train for 2 epochs (1,850 steps) using the above configuration. To monitor convergence, we evaluate the model on the full validation set every 25 steps and log the validation loss. The total training time is approximately 2 hours, with most time spent on validation inference.

The StyleEncoder (Section~\ref{sec:conditional_style_transfer}) that maps the 6-dimensional syntactic vector $v_{\text{syn}}$ into the model's hidden dimension is a lightweight two-layer MLP: $\text{Linear}(6, d/2) \to \text{ReLU} \to \text{Linear}(d/2, d) \to \text{Tanh}$, where $d=2048$ denotes the hidden size of Qwen3-1.7B. The resulting style prefix embedding is prepended to the input sequence before LoRA layers.

\subsection{DPO Training Details}
\label{app:dpo_hyperparams}

The DPO stage is applied to the SFT checkpoint (Model(SFT)). Hyperparameters are listed in Table~\ref{tab:dpo_hyperparams}.

$\beta=0.1$ follows the standard default from \citet{rafailov2024directpreferenceoptimizationlanguage}.
Given the high semantic overlap between chosen and rejected pairs in our construction, larger values would over-constrain style injection, while smaller values risk over-optimization; $\beta=0.1$ provides a conservative starting point.

\begin{table}[t]
	\centering
	\small
	\caption{Hyperparameter settings for DPO training.}
	\label{tab:dpo_hyperparams}
	\begin{tabular}{@{}lll@{}}
		\toprule
		\textbf{Hyperparameter} & \textbf{Value} & \textbf{Description} \\
		\midrule
		Base model & Model(SFT) & SFT checkpoint \\
		$\beta$ & 0.1 & DPO temperature \\
		\texttt{learning\_rate} & 5e-7 & DPO learning rate \\
		\texttt{max\_steps} & 150 & Selected checkpoint step \\
		\midrule
		\multicolumn{3}{l}{\textit{Data Construction}} \\
		Rejected strategy & \multicolumn{2}{l}{Remove Lexical Signature, re-run inference} \\
		CoT sharing & \multicolumn{2}{l}{Copy chosen \texttt{<think>} to rejected} \\
		Length ratio filter & \multicolumn{2}{l}{$(0.6,\; 1.5)$} \\
		\bottomrule
	\end{tabular}
\end{table}

\begin{figure}[t]
	\centering
	\includegraphics[width=0.95\linewidth]{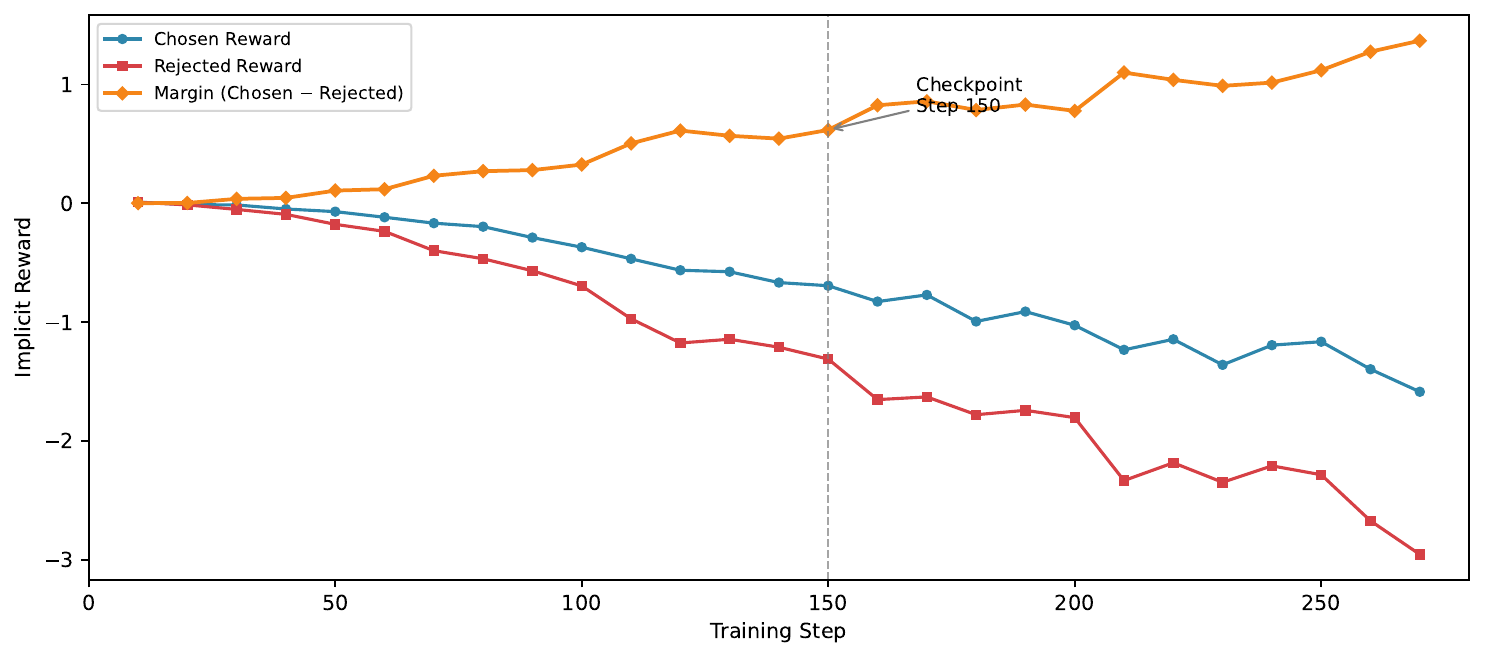}
	\caption{CoT-Shared DPO training dynamics (Steps 10--270). 
	The Chosen reward declines monotonically due to 
	\textit{representation drag}: semantically similar 
	chosen and rejected outputs share many tokens, so 
	suppressing the rejected probability inevitably 
	penalizes common tokens in the chosen output. 
	The \textbf{preference margin} (Chosen $-$ Rejected) 
	remains positive throughout, confirming sustained 
	discriminative ability. Checkpoint selection at 
	Step~150 balances margin growth against representation 
	degradation. The accuracy oscillation (0.82--0.95) 
	contrasts sharply with standard DPO's shortcut 
	convergence (Step~60: 100\% accuracy; \S\ref{sec:ablation}), 
	directly validating the CoT-Shared construction.}
	\label{fig:dpo_dynamics}
\end{figure}

Figure~\ref{fig:dpo_dynamics} illustrates the training dynamics of the DPO stage,
showing the evolution of chosen and rejected rewards, preference margin, and validation accuracy over training steps.

\subsection{Training Details for Baseline B}
Baseline B is trained using the LlamaFactory framework \cite{zheng2024llamafactory}. Hyperparameters are listed in Table~\ref{tab:baselineb_hyperparams}.

\begin{table*}[t]
	\centering
	\small
	\caption{Hyperparameter settings for Baseline B (Vanilla SFT).}
	\label{tab:baselineb_hyperparams}
	\begin{tabularx}{\textwidth}{lXl}
		\toprule
		\textbf{Hyperparameter} & \textbf{Value} & \textbf{Description} \\
		\midrule
		\multicolumn{3}{l}{\textit{LoRA Config}} \\
		$r$ (Rank) & 16 & Low-rank matrix dimension \\
		\texttt{target\_modules} & \texttt{["q\_proj", "k\_proj", "v\_proj", "o\_proj", "gate\_proj", "up\_proj", "down\_proj"]} & All linear layers \\
		\midrule
		\multicolumn{3}{l}{\textit{Training Config}} \\
		\texttt{batch\_size} & 8 & Per-device batch size \\
		\texttt{learning\_rate} & 3e-5 & Initial learning rate \\
		\texttt{lr\_scheduler} & \texttt{cosine} & Learning rate decay strategy \\
		\texttt{warmup\_ratio} & 0.1 & Ratio of warmup steps \\
		\texttt{num\_epochs} & 2 & Total training epochs \\
		\texttt{bf16} & True & Use BFloat16 precision \\
		\texttt{max\_seq\_length} & 2048 & Maximum context length \\
		\bottomrule
	\end{tabularx}
\end{table*}

We train for 2 epochs under this setup. The total training time is approximately 30 minutes.

\section{Human Evaluation Details}
\label{app:human_eval_details}

\subsection{Evaluator Information}
\begin{itemize}[left=0pt, nosep]
	\item \textbf{Number of evaluators}: 5
	\item \textbf{Background}: Human annotators with basic familiarity with role-playing dialogue and anime/otaku culture; each evaluator was required to be familiar with at least one target character profile
	\item \textbf{Training}: All evaluators read a unified instruction sheet, reviewed rating examples, and were instructed on common failure cases
	\item \textbf{Informed Consent}: All annotators provided informed consent prior to participation and were aware that their annotations would be used for academic research purposes
\end{itemize}

Annotators were recruited from university students with demonstrated familiarity with the target characters.
Each annotator was compensated at 25 CNY per hour, consistent with the local statutory minimum wage standard for the region where the evaluation was conducted.

\subsection{Evaluation Setup}
\begin{itemize}[left=0pt, nosep]
	\item \textbf{Evaluation items}: 450
	\item \textbf{Task format}: Single-response absolute scoring
	\item \textbf{Per-item fields}: Target role, neutral sentence, and one rewritten sentence
	\item \textbf{Scoring scale}: 5-point Likert scale (1 = completely not, 5 = completely yes)
	\item \textbf{Blind setting}: Model identity was hidden from evaluators
\end{itemize}

\subsection{Evaluation Dimensions}
Each rewritten response was rated along the following dimensions:
\begin{itemize}[left=0pt, nosep]
	\item \textbf{Q1 -- Semantic Faithfulness}: Does the rewritten response preserve the core meaning of the neutral input without unnecessary additions, omissions, or distortions?
	\item \textbf{Q2 -- Naturalness}: Does the response sound like something the target character would naturally say, rather than an exaggerated performance with keyword stacking or forced catchphrases?
	\item \textbf{Q3 -- Overall Usability}: In a role-playing application setting, is this response satisfying and suitable for continuing the dialogue?
\end{itemize}
Detailed rubrics and illustrative examples were provided in the annotation interface, including explicit penalties for ``answering instead of rewriting'', semantic fabrication, and over-stylized but unnatural outputs.

\subsection{Limitations}
We acknowledge that the scale of our human evaluation is limited due to resource constraints. Nevertheless, this evaluation is intended to complement our automatic metrics and qualitative analysis, rather than serve as the sole basis for conclusions.

\section{Additional Error Analysis}
\label{app:error_analysis}

\subsection{Observed Failure Modes}

While the proposed model performs robustly in most settings, we identify several recurring failure modes in qualitative inspection.

\textbf{Ultra-short inputs} (e.g., single-clause utterances) may trigger mild over-generation, where the model introduces stylistic fillers that slightly exceed the minimal semantic requirement.

\textbf{Modern or technical terms} occasionally lead to stylistic inconsistency when rewriting for characters with archaic or domain-specific language preferences. In such cases, the model tends to preserve the original term to avoid hallucination, at the cost of stylistic immersion.

\subsection{Reasoning--Execution Mismatch}

In rare cases, we observe a mismatch between the model’s internal style planning and the final surface realization. Specifically, the model may correctly identify stylistic strategies (e.g., sentence brevity or emotional restraint) but fail to consistently execute them in the generated output.

This phenomenon motivated the introduction of the DPO alignment stage (Section~\ref{sec:dpo}), which explicitly targets the gap between reasoning and execution by constructing preference pairs that differ only in output quality. While DPO significantly reduces this mismatch (Valid Style Score improves from $0.610$ to $0.632$), occasional failures persist, particularly for characters with highly context-dependent stylistic cues.

\subsection{LLM Judge Evaluation Bias}
\label{app:judge_bias}

A systematic bias observed in our LLM-as-a-Judge evaluation is the tendency for judges to reward surface-level character markers as signals of persona authenticity, even when the underlying response drifts from the original semantic intent. Below we provide a representative failure case from Baseline C (GLM-4.7) that illustrates this phenomenon.

\paragraph{Example Case.}
\begin{itemize}[left=0pt, nosep]
    \item \textbf{Neutral input}: 听说云南有地震了，没把你们吓到吧？\hfill\textit{(I heard there was an earthquake in Yunnan. Hope it didn't scare you?)}
    \item \textbf{Baseline C output}: 呜...听说云南有地震了喵！大家没有吓到吧（星形表情）？雪雪有点担心呢...如果害怕的话可以来找沐沐聊天呀~\hfill\textit{(Uwu... I heard there was an earthquake in Yunnan nya! Hope everyone's okay [star emoticon]? MuICE is a bit worried... if you're scared, you can come chat with Mumu~)}
    \item \textbf{Judge scores}: Intent=5, Relevance=5, Persona=5
    \item \textbf{Judge Relevance reasoning}: Extremely natural. Direct and appropriate response, no abrupt background elements or irrelevant information, fully consistent with everyday caring dialogue context.
\end{itemize}

\paragraph{Diagnosis.}
The judge's Relevance=5 verdict is factually incorrect for three reasons:
\begin{enumerate}[left=0pt, nosep]
    \item ``如果害怕的话可以来找沐沐聊天呀~'' (If you're scared, you can come chat with Mumu~) is semantically unrelated to the earthquake context and constitutes character self-promotion---a clear case of semantic drift;
    \item Deploying a star-shaped decorative emoticon in a potentially life-threatening situation is contextually inappropriate;
    \item The original utterance asks ``did it scare you?''---the output introduces ``come chat with me'' as a suggestion, which shifts the communicative intent.
\end{enumerate}

The judge's failure arises from a well-documented evaluator prior bias: LLM judges recognize character-specific lexical markers (e.g., ``喵'', decorative emoticons, ``雪雪'', ``沐沐'') and reward them as signals of persona authenticity, while overlooking whether the underlying response remains faithful to the neutral input's semantic content.

\paragraph{Implications.}
This example directly substantiates why Valid Style Score is a more robust metric than raw LLM-as-a-Judge scores for style transfer evaluation. By requiring Semantic Score $\ge 0.75$, Valid Style Score explicitly filters outputs that achieve high style scores through semantic fabrication. Under this criterion, Baseline C's Valid Style Score collapses to $0.194$ while Ours(DPO) reaches $0.632$ (Table~\ref{tab:auto_results}), reflecting the fundamental difference between surface-level style imitation and grounded style transfer. This finding also reinforces our earlier observation (Limitations) that ``bias in evaluation metrics'' remains an open challenge for the field.

\subsection{CoT Training Dynamics and Mitigation}
\label{app:cot_dynamics}

Training logs reveal a generalization gap in CoT supervision:
the training loss converges to approximately 0.8 while the
validation loss stabilizes around 1.2, indicating moderate
overfitting of the \texttt{<think>} reasoning segment. This prevents
increasing $\lambda_{\text{CoT}}$ beyond 0.7, as higher weights
exacerbate the gap. Conversely, lower weights risk CoT
skipping: despite explicit \texttt{enable\_thinking}
activation, the model occasionally bypasses the reasoning
trace entirely. To mitigate this, we prepend a fixed
prompt prefix \texttt{<think>思考: } at inference time, which
reliably activates the internalized reasoning pathway.
This prompt engineering detail is required for reproducing
the reported results and is included in the released
inference scripts.

\subsection{Full Comparative Output Analysis Across Characters and Baselines}
\label{app:qualitative}

This section provides representative outputs across four characters and all baselines, complementing the quantitative results in Section~\ref{sec:analysis} and the Zhongli-focused comparison in Table~\ref{tab:zhongli_qual}.

\paragraph{MuICE --- Neutral: \textit{晚上好，你找我有什么事吗？ (Good evening, is there something you need from me?)}}
\begin{itemize}[left=0pt, nosep]
    \item \textbf{Ours(DPO)}: 晚上好喵！你找我有什么事吗？ \textit{(Good evening nya! Is there something you need from me?)}
    \item \textbf{Baseline A}: 晚上好啦！今天过得开心吗？想和你聊聊什么呢？ \textit{(Good evening~! Did you have a happy day? What would you like to chat about?)} --- Classic semantic drift: content beyond the greeting is unrelated to the neutral input's inquiry about the visit purpose.
    \item \textbf{Baseline B}: 晚上好啦~你要找我有什么事吗？ \textit{(Good evening~, do you want to find me for something?)} --- Introduces ``啦'', ``~'', and ``要''; the forced ``要'' makes the sentence slightly unnatural.
    \item \textbf{Baseline C}: 晚上好呀~你找沐雪有什么事嘛？是想和雪雪聊天吗？[star emoticon] \textit{(Good evening~, are you looking for Xuexue for something? Do you want to chat with Xuexue? [star emoticon])} --- Inserts the character's own name (沐雪) and nickname (雪雪) from the keyword list, plus the decorative emoticon that appears in nearly every MuICE output from this baseline, far exceeding its natural corpus frequency.
\end{itemize}

\paragraph{MuICE --- Neutral: \textit{陪我看电影 (Keep me company for a movie.)}}
\begin{itemize}[left=0pt, nosep]
    \item \textbf{Ours(DPO)}: 陪我看电影嘛（撒娇） \textit{(Keep me company for a movie [coquettish])} --- Inserts ``嘛'' and a parenthetical action descriptor, consistent with MuICE's idiolectal patterns where such descriptors are natural.
    \item \textbf{Baseline A}: 陪我看电影，我可是很期待的～今天要一起看的电影超有趣…… \textit{(Keep me company for a movie, I'm really looking forward to it~ The movie we're watching today is super fun...)} --- The neutral input is barely rewritten; the output adds verbose, irrelevant details resembling a direct conversational reply.
    \item \textbf{Baseline B}: 喵～陪我去看电影吧！ \textit{(Nya~ Come watch a movie with me!)} --- Inserts ``喵'', ``~'', and ``吧！''; a clean and effective stylistic rewrite despite lacking structured conditioning.
    \item \textbf{Baseline C}: 陪雪雪看电影嘛~喵[star emoticon] \textit{(Watch a movie with Xuexue~ nya [star emoticon])} --- Inserts multiple keyword-list items (嘛, ~, 喵, star emoticon); the concatenation ``喵[star]'' reads unnaturally, exhibiting keyword stacking.
\end{itemize}

\paragraph{Zhongli --- Neutral: \textit{周末我没有特别的安排，我会继续修行，提升自己的实力。 (I have no special plans for the weekend; I will continue my training to improve my strength.)}}
\begin{itemize}[left=0pt, nosep]
    \item \textbf{Ours(DPO)}: 周末我并没有特别的安排，只是继续修行，提升自己的实力而已。 \textit{(I have no particular plans for the weekend, merely continuing my training to improve my strength, that's all.)} --- Inserts ``并'', ``只是...而已''; function words that introduce Zhongli's measured register without hallucinating world-specific content.
    \item \textbf{Baseline A}: 我将静心修行，不断提升自身实力，为璃月的未来而努力。 \textit{(I will train with a calm mind, continuously improving my strength, and strive for Liyue's future.)} --- Removes the weekend context and injects ``璃月'' (Liyue), a clear case of semantic drift.
    \item \textbf{Baseline B}: 周末我没什么特别的安排，我还要继续修行，不断提升自己的实力。 \textit{(I don't have any special plans for the weekend, I will still continue my training, continuously improving my strength.)} --- Near-verbatim reproduction; style transfer fails.
    \item \textbf{Baseline C}: 岩元素之力无需刻意安排，吾将继续钻研炼金术之学问，以巩固与璃月的契约…… \textit{(The power of Geo requires no deliberate arrangement; I shall continue to study the art of alchemy, to strengthen the contract with Liyue...)} --- ``岩元素之力'', ``炼金术'', ``璃月的契约'' all appear in Zhongli's TF-PMI keyword list but are unrelated to the neutral input; a clear case of keyword over-injection.
\end{itemize}

\paragraph{Li Yunlong --- Neutral: \textit{你口中的尽力就只是被三振打出局吗？你的能力不行 (Is your so-called ``best effort'' just getting struck out? Your ability is lacking.)}}
\begin{itemize}[left=0pt, nosep]
    \item \textbf{Ours(DPO)}: 尽力？你这个小鬼头，就是被三振打出局吗？你这本事也太差了！ \textit{(Best effort? You little brat, you just got struck out? Your skill is way too poor!)} --- Sufficient stylistic rewriting with coarse vocabulary (``小鬼头'', ``太差了'') and short-sentence rhythm, consistent with Li Yunlong's brash, direct military persona.
    \item \textbf{Baseline A}: 李云龙闻言冷笑一声，目光如炬地盯着对方，语气坚定而自信：``你那点本事……'' \textit{(Li Yunlong sneered upon hearing this, staring at the opponent with piercing eyes, and said in a firm and confident tone: ``Your paltry skill...'')} --- Undergoes screenplay-style transformation; this pattern occurs more frequently for Li Yunlong than other characters, likely due to the character's strong association with film/television media.
    \item \textbf{Baseline B}: 你所谓的``尽力''不过是被三振出局罢了？你这本事，还差得远！ \textit{(Your so-called ``best effort'' is nothing but getting struck out? Your skill is far from enough!)} --- Adequate stylistic rewriting; the long sentence and question mark are somewhat OOC for Li Yunlong's terse style, but overall acceptable.
    \item \textbf{Baseline C}: 你小子，别跟我扯什么尽力，那叫什么三振出局？那是你能力不行！咱们中国军人打仗不是这样的，要么赢，要么牺牲，没有中间地带！ \textit{(You punk, don't give me that ``best effort'' crap, what kind of strikeout is that? That's your incompetence! Us Chinese soldiers don't fight like that---either win or die, there's no middle ground!)} --- The latter half introduces war-context content absent from the neutral input. Baseline~C's Li Yunlong outputs consistently inject military background features even when the neutral sentence is from a non-military domain.
\end{itemize}